\documentclass[journal]{IEEEtran}
\usepackage{cite}

\ifCLASSINFOpdf
\else
\fi
\usepackage{amsmath}
\usepackage{algorithmic}

\usepackage{array}

\ifCLASSOPTIONcompsoc
  \usepackage[caption=false,font=normalsize,labelfont=sf,textfont=sf]{subfig}
\else
  \usepackage[caption=false,font=footnotesize]{subfig}
\fi

\usepackage{stfloats}

\usepackage{url}

\usepackage{gensymb}
\usepackage{graphicx}
\usepackage{tabulary}
\usepackage{siunitx}
\usepackage{caption}
\newcommand\T{\rule{0pt}{2.6ex}}       
\newcommand\B{\rule[-1.2ex]{0pt}{0pt}} 
\usepackage{ragged2e}
\usepackage{textcomp}

\usepackage[pdftex,implicit=false,colorlinks=true,urlcolor=blue]{hyperref}

\begin{document}
%
\title{DOTS: An Open Testbed for Industrial Swarm Robotic Solutions}
%
%
%
\author{Simon~Jones, Emma~Milner, Mahesh~Sooriyabandara, and Sabine Hauert
\thanks{S. Jones, E. Milner, and S. Hauert are at the University of Bristol, UK. simon2.jones@bristol.ac.uk, emma.milner@bristol.ac.uk, sabine.hauert@bristol.ac.uk}
\thanks{M. Sooriyabandara is at Toshiba Research, Bristol, UK, Mahesh.Sooriyabandara@toshiba-bril.com}
}
\maketitle

\begin{abstract}
We present DOTS, a new open access testbed for industrial swarm robotics experimentation. It consists of 20 fast agile robots with high sensing and computational performance, and real-world payload capability. They are housed in an arena equipped with private 5G, motion capture, multiple cameras, and openly accessible via an online portal. We reduce barriers to entry by providing a complete platform-agnostic pipeline to develop, simulate, and deploy experimental applications to the swarm. We showcase the testbed capabilities with a swarm logistics application, autonomously and reliably searching for and retrieving multiple cargo carriers.
\end{abstract}


%
\IEEEpeerreviewmaketitle




\section{Introduction}
%
%
%
%
\IEEEPARstart{M}{any} robots (10-1000+) working together to facilitate intralogistics in real-world settings promise to improve productivity through the automatic transport, storage, inspection, and retrieval of goods. Yet, existing systems typically require sophisticated robots, carefully engineered infrastructure to support the operations of the robots, and often central planners to coordinate the many-robot system (e.g. Amazon, Ocado). Once operational, these solutions offer speed and precision, but the pre-operational investment cost is often high and the flexibility can be low. 

Consequently, there is a critical unmet need in scenarios that would benefit from logistics solutions that are low-cost and usable out-of-the-box. These robot solutions must adapt to evolving user requirements, scale to meet varying demand, and be robust to the messiness of real-world deployments. Examples include small and medium enterprises, local retail shops, pop-up or flexible warehouses (COVID-19 distribution centres, food banks, refugee camps, airport luggage storage), and manufacturing of products with high variance or variation (e.g. small series, personalised manufacturing). Flexible solutions for such scenarios also have the potential to translate to other applications and settings such as construction or inspection.

Swarm robotics offers a solution to this unmet need. Large numbers of robots, following distributed rules that react to local interactions with other robots or local perception of their environment, can give rise to efficient, flexible, and coordinated behaviours. A swarm engineering approach has the potential to be flexible to the number of robots involved, enabling easy scalability, adaptation and robustness. While traditional logistics solutions prioritise throughput, speed and operating cost, swarm logistics solutions will be measured against additional ‘Swarm Ideals’, targeting zero setup, reconfiguration time, and scaling effort, zero infrastructure, zero training, and zero failure modes. It is these attributes that make swarm robotics the natural solution for logistics in more dynamic and varied scenarios.


\begin{figure}
  	\centering
  	\includegraphics[width=0.95\columnwidth]{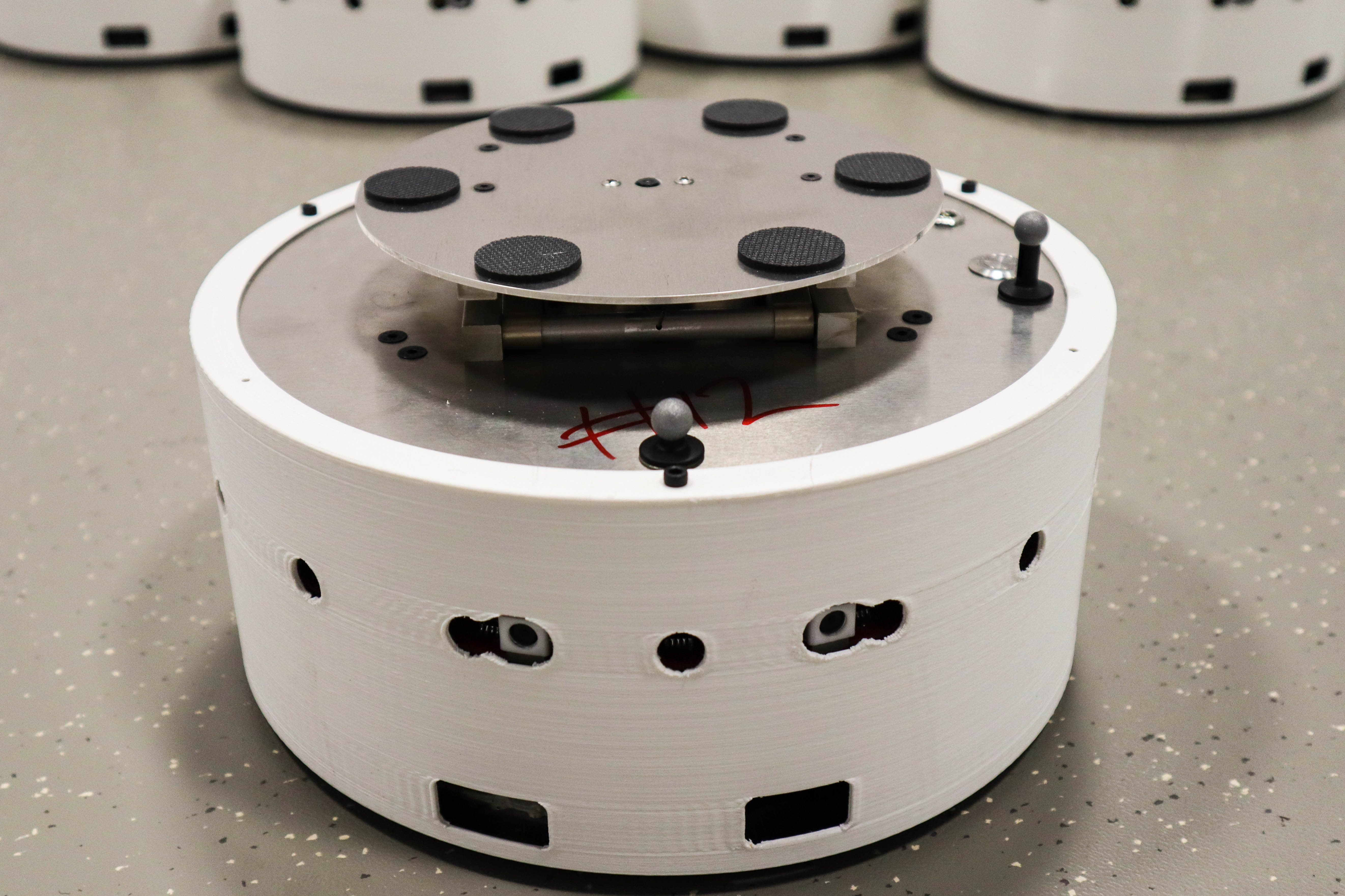}
  	\caption{DOTS robot, showing two forward-facing cameras and lifting platform on top. }\label{fig:dotsrobot}
\end{figure}

To explore this potential, we present a new 5G-enabled testbed for industrial swarm robotic solutions. The testbed hosts 20 custom-built \SI{250}{mm} robots called DOTS (Distributed Organisation and Transport System) that move fast, have long battery life (6 hours), are 5G enabled, house a GPU, can sense the environment locally with cameras and distance sensors, as well as lift and transport payloads (\SI{2}{kg} per robot). The platform is modular and so can easily be augmented with new capabilities based on the scenarios to be explored (e.g. to manipulate items). To monitor experiments, the arena is fitted with overhead cameras and a motion capture system, allowing for precise telemetry and replay capabilities. The testbed is accessible remotely through custom-built cloud infrastructure, allowing for experiments to be run from anywhere in the world and enabling future fast integration between the digital and physical word through digital twinning. In addition we present an integrated development environment useable on Windows, Linux and OSX lowering the barriers to entry for users of the system. 

\begin{figure*}
  	\centering
  	\includegraphics[width=0.8\textwidth]{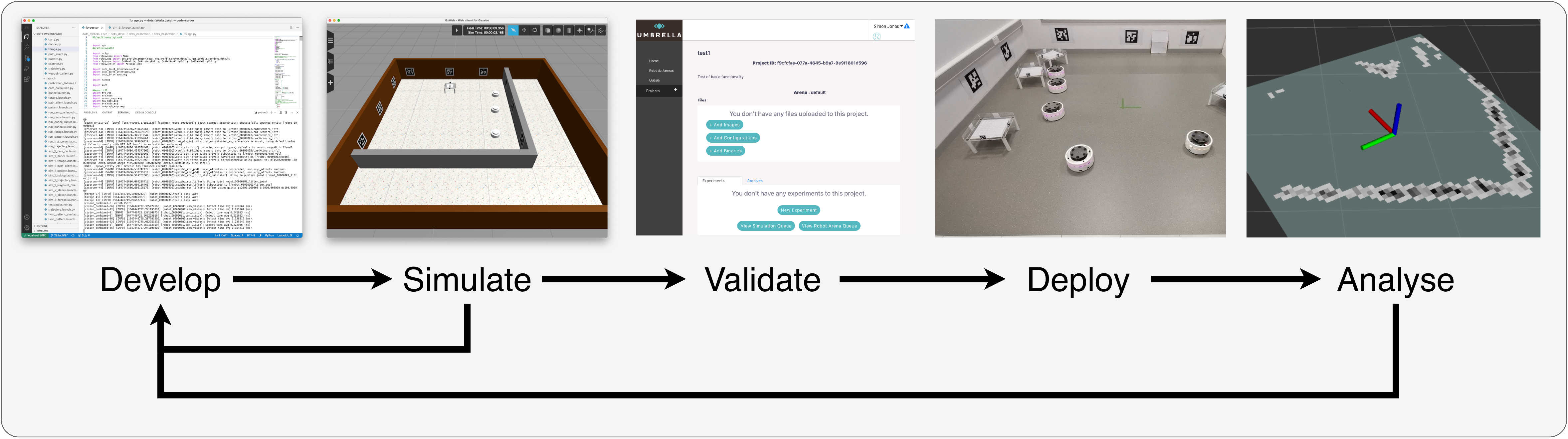}
  	\caption{Using the DOTS IDE, controllers are developed and tested locally in simulation. When ready, controllers are validated using the online portal and deployed to the physical testbed for an experimental run. Experimental data captured is available for download from the portal and subsequent analysis.}\label{fig:pipeline}
\end{figure*}

Beyond characterising the testbed, making use of the development pipeline illustrated in Figure \ref{fig:pipeline}, we demonstrate the steps and processes required to take a conceptually simple swarm logistics algorithm and successfully run it on real-world robots to reliably find, collect, and deposit five payload carriers in an entirely decentralised way.

This paper is structured in the following way; in Section \ref{sec:rel} we provide background and discuss related work. In Section \ref{sec:methods} we detail the physical, electronic, and software systems of the testbed, with characterisation of the performance of various subsystems. In Section \ref{sec:demo} we build a complete demonstration of a distributed logistics task. Section \ref{sec:concs} concludes the article.

\section{Background and related work}\label{sec:rel}
 Two recent reviews of multi-robot technologies for warehouse automation show that amongst over 100 papers in the area, only very few considered decentralised, distributed, or swarm approaches \cite{custodio2020flexible,jaghbeer2020automated}. Of those that did, solutions were only partially decentralised \cite{draganjac2016decentralized}. This suggests  swarm solutions for logistics have largely remained unexplored, although the interest in distributed multi-robot systems for intralogistics is growing as shown in a recent review by \cite{fragapane2021planning}. At the same time, swarm robotics \cite{csahin2005swarm} has 30 years of history, mostly focussing on conceptual problems in the laboratory with translation to applications increasing in the last couple of years \cite{schranz2021swarm}. 
 
 We now need fresh strategies to design and deploy swarm solutions that eschew the mantra of emergence from interaction of simple agents and embrace the use of high specification robots at the agent level to enhance the performance of a robot swarm without sacrificing its inherent scalability and adaptability in real-world applications. This is made possible now due to the convergence of technologies including high individual robot specifications (fast motion, high-computation, and high-precision sensing), 5G networking to power communication between humans and interconnected robots, access to high onboard computational power that allows for sophisticated local perception of the world and new algorithms for swarm control. Swarm testbeds for research do exist. 
 The Robotarium \cite{pickem2017robotarium} is a complete system of small robots, with associated simulator, online access, tracking, and automated charging and management. The individual robots are not autonomous though, with controller code executes on a central server, with the robots acting as peripherals. Duckietown \cite{paull2017duckietown} is an open source platform of small cheap autonomous robots designed for teaching autonomous self-driving, but the emphasis is on users building their own testbed. The Fraunhofer IML Loadrunner \cite{ten2020technical} swarm is technically sophisticated and physically considerably larger than our design but does not appear to be open.

\section{Materials and Methods}\label{sec:methods}
In the following sections, we detail the robots, the arena, the online portal for remote experimentation, and the cross-platform integrated development environment.

\subsection{DOTS robots}
The robots were designed from the start to be low-cost, around \pounds 1000 per robot, simple to construct, to allow relatively large numbers to be built, and high capability, by using commodity parts in innovative ways. Each robot is \SI{250}{mm} in diameter with a holonomic drive capable of rapid omnidirectional movement, up to \SI{2}{ms$^{-1}$} and typically 6 hours of battery life. They are equipped with multiple sensors; \SI{360}{\degree} vision with four cameras and a further camera looking vertically upwards, laser time-of-flight sensors for accurate distance sensing of surrounding obstacles, and multiple environment sensors. Each robot has a lifting platform, allowing the transport of payloads. Onboard computation is provided with a high-specification Single Board Computer (SBC), with six ARM CPUs and a GPU. And there are multiple forms of communication available - WiFi, two Bluetooth Low Energy (BLE5) programmable radios, an ultrawideband (UWB) distance ranging radio, and the ability to add a 5G modem.

\begin{table}\centering\caption{Features of the DOTS robots\B}\label{tab:specs}
{\RaggedRight
\begin{tabulary}{\columnwidth}{p{0.25\columnwidth}lp{0.38\columnwidth}}
	\hline
	Specification & Value & Notes \T\B \\
	\hline
Diameter & 250 mm &  \\
Height & 145 mm & Lifter lowered \\
& 197mm & Lifter raised \\
Weight & 3 kg &\\
Lifter max load & 2 kg &\\
Max velocity & 2 ms$^{-1}$ &\\
Max acceleration & 4 ms$^{-2}$ &\\
Wheel torque & 0.35 Nm &\\
Battery capacity & 100 Wh & 2x Toshiba SCiB \SI{23}{Ah}\\
Endurance & 6 hours & All cameras enabled, moderate vision processing and movement\\
& 14 hours & No cameras, low level processing, occasional movement \\
Carriers & 330x\SI{330}{mm}  & \SI{175}{mm} clearance \\
\hline
Processor & RockPi4 & 6x ARM CPUs, ARM Mali T860MP4 GPU \\
Cameras & OV5647 & 5x 120\degree FOV video up to 1080p30 \\
Proximity & VL53L1CX & 16x IR laser time-of-flight, \SI{3}{m} range, \SI{10}{mm} precision \\
IMU & LSM6DSM & 6DoF accelerometer and gyroscope\\
Magnetometer & LIS2MDL & 3DoF\\
Temperature, humidity& Si7021-A20 & Temperature accuracy $\pm$\SI{0.4}{\degree C}, relative humidity $\pm$3\% \\
Absolute pressure & LPS22HB & \SI{260}{hPa}- \SI{1260}{hPa}, accuracy $\pm$\SI{0.1}{hPa}\\
Microphones &IM69D130& MEMS sensor, FPGA-based CIC filter\\
 \hline
\end{tabulary}
}
\end{table}

\begin{figure}
  	\centering
  	\includegraphics[width=1.0\columnwidth]{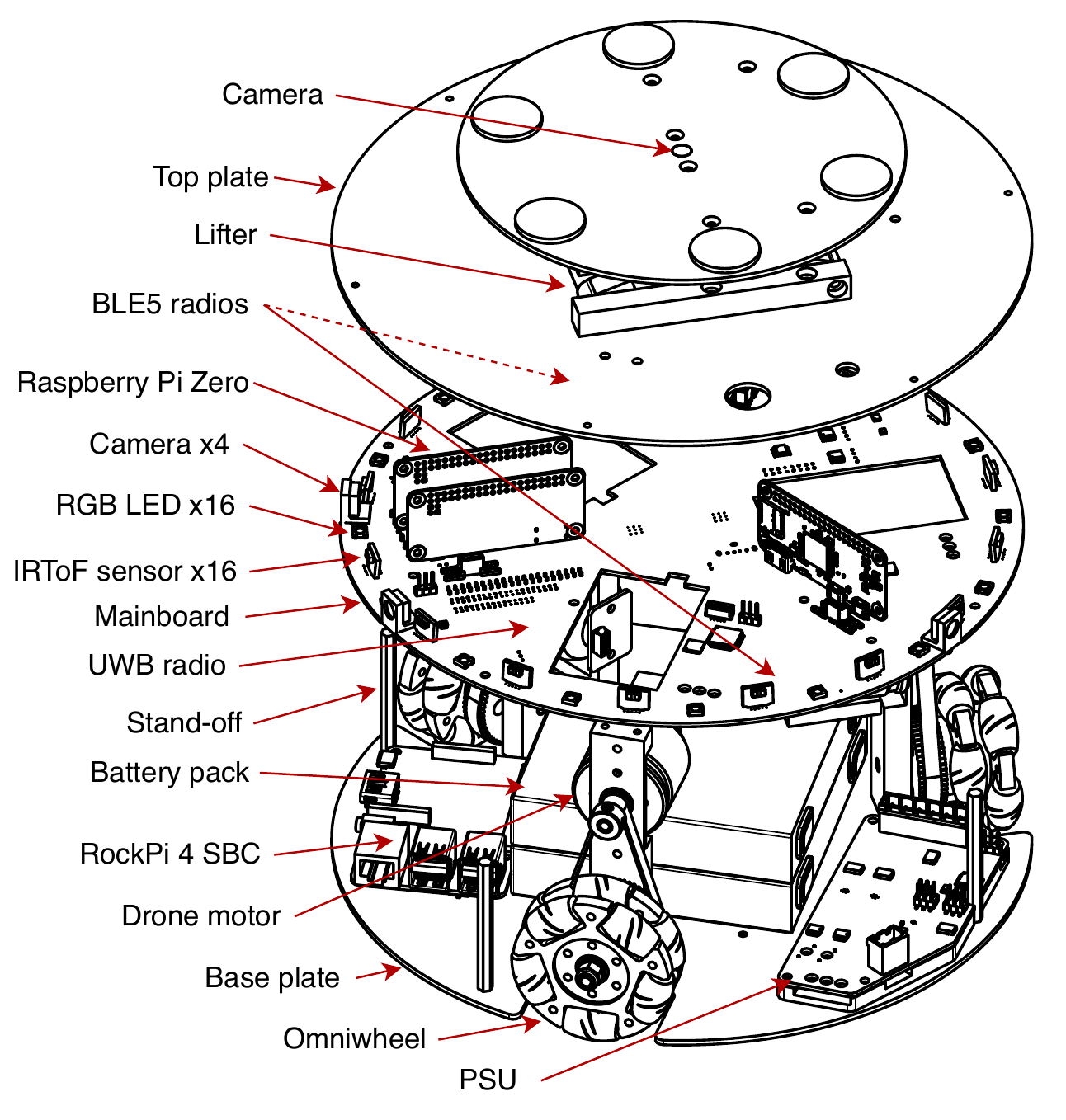}
  	\caption{Exploded view of the robot chassis showing major components.}\label{fig:robotexpl}
\end{figure}

\subsubsection{Cost minimisation}
A central driving factor in the design process was cost minimisation while still achieving good performance. The rapid progress in mobile phone capabilities means that high performance sensors such as cameras are now available at very low cost. Communications advances means that fully programmable Bluetooth Low Energy (BLE5) modules cost only a few pounds. Single board computers based on mobile phone SoCs are widely available, high definition cameras cost \pounds4. And the market for consumer drones has made very high power-to-weight ratio motors available at much lower cost than specialised servo motors. We leverage these advances to reduce the  cost per robot to \pounds 1000, so the complete testbed can have many robots.

\subsubsection{Robot chassis}
The mechanical chassis, shown in Figure \ref{fig:robotexpl}, is designed to be easy to fabricate in volumes of a few 10s. Custom parts are made using 3D printing or simple milling operations. It is constructed around two disks of \SI{2}{mm} thick aluminium, \SI{250}{mm} in diameter, for the base and top surfaces. Between the base and top plates and mechanically joining them \SI{100}{mm} apart are three omniwheel assemblies, positioned at \SI{120}{\degree} separation. Each wheel assembly consists of an omniwheel, a drone motor with rotational position encoder, and a 5:1 timing belt reduction drive between motor and wheel. Mounted on the base plate are the single board computer (SBC) and the power supply PCBs, in both cases thermally bonded to the base plate to provide passive cooling. The battery pack sits in the central area of the base plate. On \SI{55}{mm} standoffs above the base plate is the mainboard PCB, which contains all the sensors, four cameras, motor drives, and associated support electronics. The top plate holds the payload lifting mechanism and an upward-facing camera, connecting via USB and power leads to the mainboard. Alternate top plates, with different actuators, for example a small robot arm, can easily be fitted.

\subsubsection{System architecture}
\begin{figure}
  	\centering
  	\includegraphics[width=0.95\columnwidth]{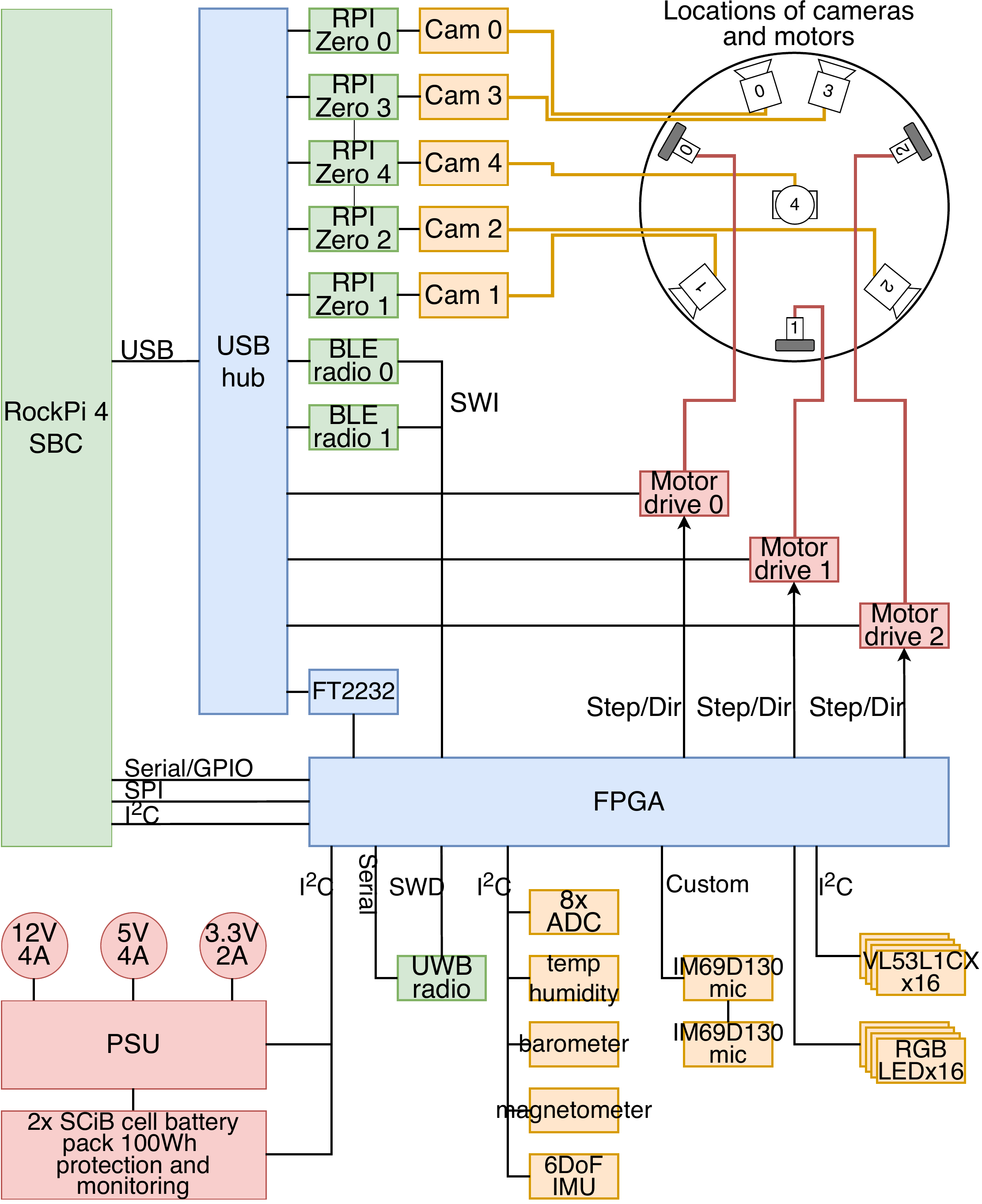}
  	\caption{Architecture of the robot. Green for computation and communication, red for power, orange for sensors, and blue for interconnect fabric.}\label{fig:robotarch}
\end{figure}

Many subsystems within the robot need to be integrated and connected to the single board computer. The system architecture is shown in Figure \ref{fig:robotarch}. Communications between sensors, actuators, and the single board computer takes place on three bus types; USB for high bandwidth, mostly MJPEG compressed video from the cameras, I$^2$C for low bandwidth, mostly various sensors, and SPI for medium bandwidth deterministic latency, used for controlling the motor movements. An FPGA is used for glue logic, for example to interface to the MEMS microphones and the programmable LEDs. The BLE5 and UWB radios are reprogrammable from the SBC via their single wire debug (SWD) interfaces \cite{williams2009low}.

\subsubsection{Vision}
Vision is the primary modality by which the robot can understand its environment. Miniature cameras have been revolutionised by progress in mobile devices. It is now possible to buy a 5 megapixel camera also capable of multiple video resolutions up to 1080p for \pounds 4. A central design issue is how to best use this cheap data - image processing is computationally heavy. We designed the system to have all round vision for the robot, achieved by using four cameras with a \SI{120}{\degree} field of view (FOV) around the perimeter of the robot. The two front-facing cameras have overlapping fields of view to allow for the possibility of stereo depth extraction. A fifth camera is fitted in the centre of the lifting platform looking upwards so that visual navigation under load carriers can be achieved with suitable targeting patterns.


The amount of data in an uncompressed video stream is high, 640x480 at \SI{30}{fps} takes around \SI{18}{MBytes/s} of bandwidth. For five cameras, this is close to \SI{100}{MBytes/s}, which would require a high performance interconnect such a gigabit ethernet, impractical at low cost and indicating a need for image compression. Rather than using fixed function webcam hardware to perform this compression, we looked for flexibility. The approach we took was to use a local Raspberry Pi Zero computer for each camera to perform image compression and then send the data over USB. This gives a cost per camera of around \pounds 15.

The Raspberry Pi Zero is a low cost small form factor Single Board Computer with a camera interface. Although the CPU has quite a low performance (ARM1176 700MHz), it is not generally appreciated how powerful the associated VPU (Video Processing Unit) and GPU (Graphics Processing Unit) is. This allows for the possibility of utilising this power to perform local image processing, e.g. to perform fiducial recognition locally, rather than loading the main central processor. In this paper, we use standard applications to stream MJPEG compressed frames with low latency from the photon arrival at the camera, to the presence of data in a buffer at the central processor.


Although cameras are cheap, processing image streams to give useful information is not, and although the single board computer is quite capable, it is not in the same class as a desktop PC. This means we are not free to easily run more complex algorithms without considerable optimisation efforts to, for example, utilise GPU processing. As an example, most visual SLAM algorithms are evaluated on PC-class systems. ORB-SLAM2 was evaluated in a system with an Intel Core i7-4790 and this is just fast enough to run at real-time with a single camera at 640x480 \SI{30}{Hz} \cite{mur2017orb}. With this in mind, we can ease the task of understanding the environment by using ubiquitous fiducial markers - future systems will be able to use greater processing power. The ArUco library \cite{romero2018speeded} has a far lower computational load than visual SLAM, and this allows us to process all five camera streams and extract marker poses from each.

\paragraph{Vision latency}
A basic performance metric when using vision for motion control is the time delay between photons arriving at the camera, and when data corresponding to those photons is available in a memory buffer on the single board computer, known as \textit{Glass-to-Algorithm} \cite{bachhuber2017minimization}. Measuring the vision latency is non-trivial, see \cite{bachhuber2017today}. We used a set of eight LEDs in approximately the centre of the field of view of one camera, and performed image processing on the destination buffer to extract the binary data displayed on the LEDs. By setting a code on the LEDs then timing how long before the code was visible in the buffer data we can measure the latency. After each code is detected, the next in a Gray-code sequence is displayed to accumulate multiple timings. 

Data from each camera is transmitted to the single board computer via a USB hub on the mainboard. We load the system by running a varying number of cameras simultaneously. Measurements are taken with different camera refresh rates. The results are shown in Table \ref{tab:visionlatency} for different refresh rates with varying numbers of other cameras running and streaming data at the same time. Times averaged over 1000 samples.

\begin{table}\centering\caption{Vision system latency, from photon arrival to change in memory buffer (\textit{glass-to-algorithm}) under different camera refresh rates and utilisation of the USB fabric. Results are averaged over 1000 frames.}\label{tab:visionlatency}
\begin{tabulary}{\textwidth}{CCCC}
	\hline
	Cameras & Latency $\bar{x}$ (ms) & Latency $\sigma$ (ms) & Latency (frames)\\
	\hline
	\multicolumn{3}{l}{\SI{30}{Hz} refresh rate} \\
1 & 78.2 & 9.23 & 2.35\\
2 & 72.8 & 10.7 & 2.18\\
3 & 74.6 & 11.7 & 2.24\\
4 & 73.4 & 11.5 & 2.20\\
5 & 78.5 & 11.5 & 2.36\\
	\multicolumn{3}{l}{\SI{60}{Hz} refresh rate} \\
1 & 53.2 & 15.8 & 3.19\\
2 & 45.2 & 9.55 & 2.71\\
3 & 45.3 & 9.82 & 2.72\\
4 & 58.3 & 12.9 & 3.50\\
5 & 58.4 & 14.6 & 3.50\\
	\hline
\end{tabulary}
\end{table}

Because the LEDs are positioned at approximately the centre vertically of the camera field of view, and the camera performs a raster scan to read the sensed image, there is a minimum of half a frame before data is available to the Raspberry Pi for processing and compression. Since the LEDs are changed uncorrelated to the frame rate, there may be between zero and one additional frame of delay uniformly distributed, giving a mean of one frame delay before the start of processing, compression, transmission, and then decompression into the destination image buffer.

The worst-case figures of \SI{79}{ms} at \SI{30}{Hz} and \SI{58}{ms} at \SI{60}{Hz} for the most heavily loaded cases of five cameras streaming image data compare well with state-of-the-art systems, which is reported as 50-\SI{80}{ms} by \cite{bachhuber2017minimization}.

\paragraph{Camera calibration}
In order that the perimeter cameras can be used to extract pose information from fiducial markers, it is necessary that they be calibrated so that their intrinsic parameters are known. We initially calibrated several cameras using the ArUco calibration target and software tools. The process involves taking multiple images of the calibration target then using the software tool to find intrinsics that minimise the reprojection error. This showed that each camera had significant differences in their intrinsic parameters, perhaps not surprising given their very low cost, but meaning we had to calibrate each camera individually, rather than using a standard calibration. We needed an automated approach - 20 robots, each with four cameras to calibrate would require many hundreds of images captured from different angles. 

By attaching a calibration target to the arena wall at a fixed and known location, we could execute a calibration process automatically taking multiple pictures per camera in about 5 minutes per robot. Because the only angle that can be varied between the camera and the target is the rotation about robot Z, the calibration is less complete, but sufficient given the robot will always be horizontal on the arena floor.

\begin{figure}
  	\centering
  	\includegraphics[width=0.98\columnwidth]{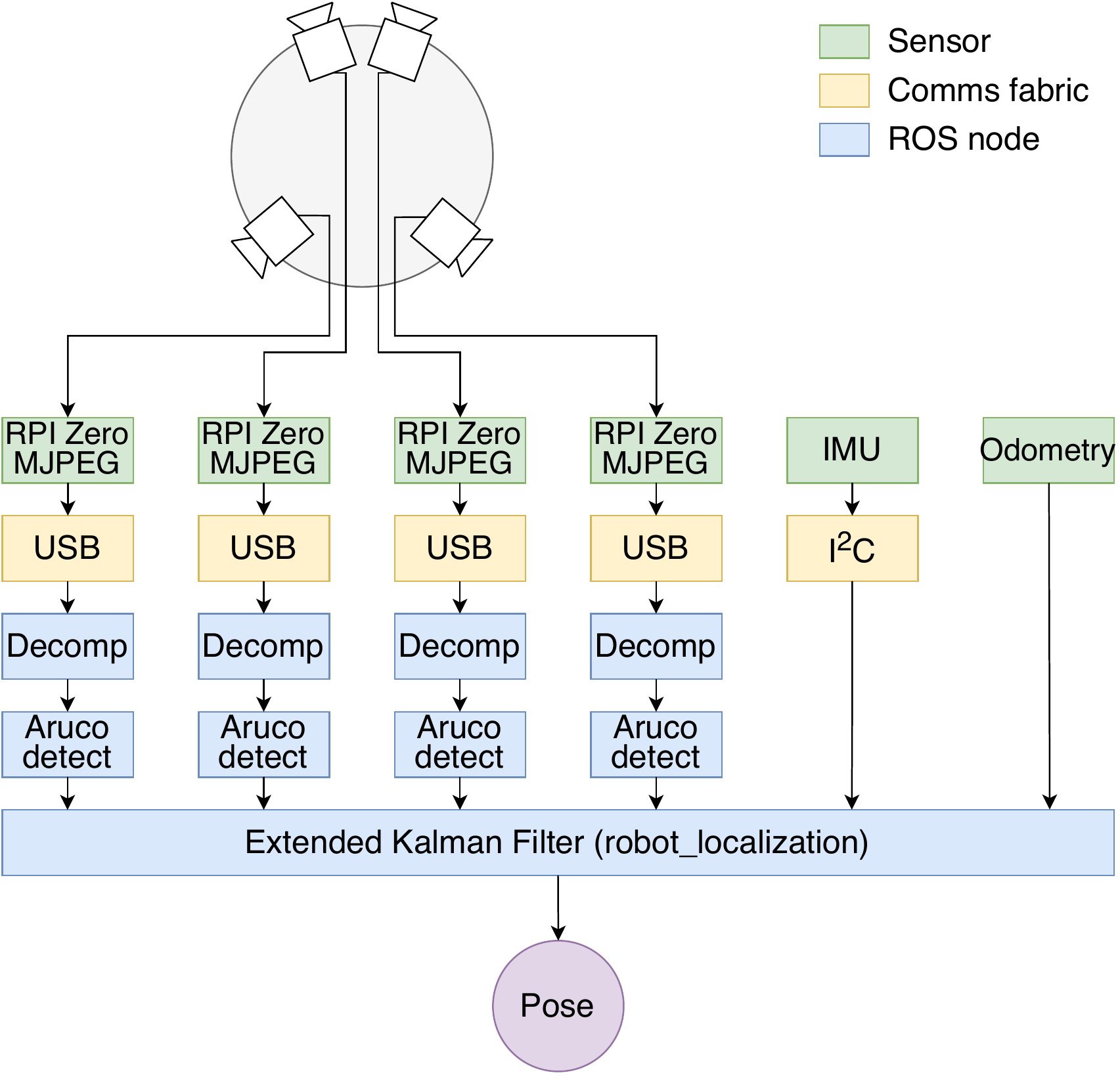}
  	\caption{Visual tracking system. Each camera video stream is send as compressed MJPEG to the single board computer, then decompressed and processed for fiducial marker poses. The poses, together with IMU and wheel odometry information are fed to an EKF to output robot pose.}\label{fig:vistracksys}
\end{figure}

\begin{figure}
  	\centering
  	\includegraphics[trim=15 0 0 35,width=1.02\columnwidth]{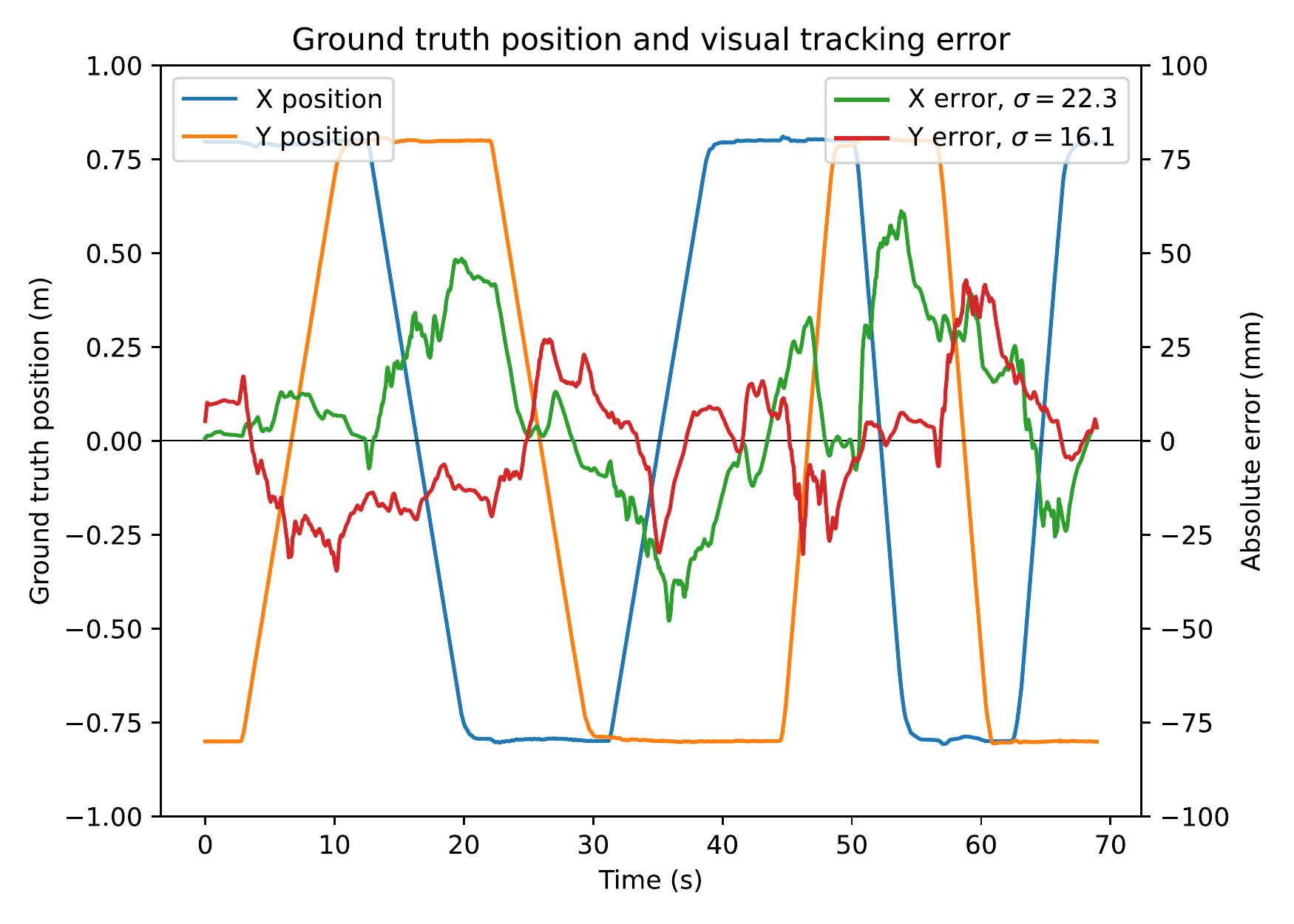}
  	\caption{Ground truth X and Y positions (left axis), and visual tracking position error (right axis), of robot while moving along a square path.}\label{fig:vistrack}
\end{figure}

\paragraph{Vision processing for localisation}\label{sec:visloc}
To demonstrate the ability to usefully process vision from multiple cameras we built a simple localisation system, shown in Figure \ref{fig:vistracksys}, and based on ArUco markermaps. This is a feature of the ArUco library that allows the specification of a map of the locations of an arbitrary number of arbitrarily placed fiducial markers, with library functions to return the 3D pose of a calibrated camera which can see at least one of the markers in the map. We fixed twelve fiducial markers around the walls of the arena, and encoded the locations in a markermap.

Each camera was calibrated automatically, as described above. The video stream from each of the four perimeter cameras is analysed and if there are any visible fiducials in the correct ID range, the pose of the robot in the global \textit{map} frame is generated. This stream of poses is fed to an Extended Kalman Filter  filter (EKF, \textit{robot\_localization} \cite{moore2014ageneralized}), along with IMU and wheel odometry information. The output of the EKF is a stream of poses in the \textit{map} frame. 

The robot was commanded to move twice around a square of  sides 1.6m, with velocities of 0.3ms$^{-1}$ and 0.5ms$^{-1}$ using ground truth as the position feedback. Ground truth from the motion capture system and estimated position were recorded. Figure \ref{fig:vistrack} shows the actual positions and the absolute error in $x$ and $y$ axes. Maximum positional error in either axis never exceeds 62mm, with $\sigma(x)=22.3$mm and $\sigma(y)=16.1$mm.

\subsubsection{Sensors}
In addition to the vision system, the are a wide variety of sensors to apprehend the state of the environment and the robot itself.

\paragraph{Proximity sensors}
Surrounding the perimeter of the robot are 16 equally-spaced ST Microelectronics VL53L1CX infra-red laser time-of-flight distance sensors (IRToF). These are capable of measuring distance to several metres with a precision of around \SI{10}{mm} at an update rate of \SI{50}{Hz}, giving a pointcloud of the environment around the robot. Each detector has a field of view of approximately \SI{22}{\degree}, which can be partitioned into smaller regions, allowing for a higher resolution pointcloud at the cost of lower update rate. 


As well as returning distance measurements, the VL53L1CX devices are capable of being operated in a mode that returns raw photon counts per temporal bin. This opens up intriguing possibilities - it is possible to classify materials by their temporal response to incident illumination, and this is demonstrated in \cite{su2016material} with a custom built time-of-flight camera. \cite{callenberg2021low} show that this also possible with the VL53L1CX device, despite the much lower cost and limitations on output power and temporal resolution. They demonstrate successful identification of  five different materials. There is no reason in principle that this robot could not function as a mobile material identification platform, for example in inspection applications.

\begin{figure*}
  	\centering
  	\includegraphics[trim=0 15 0 30,width=0.75\textwidth]{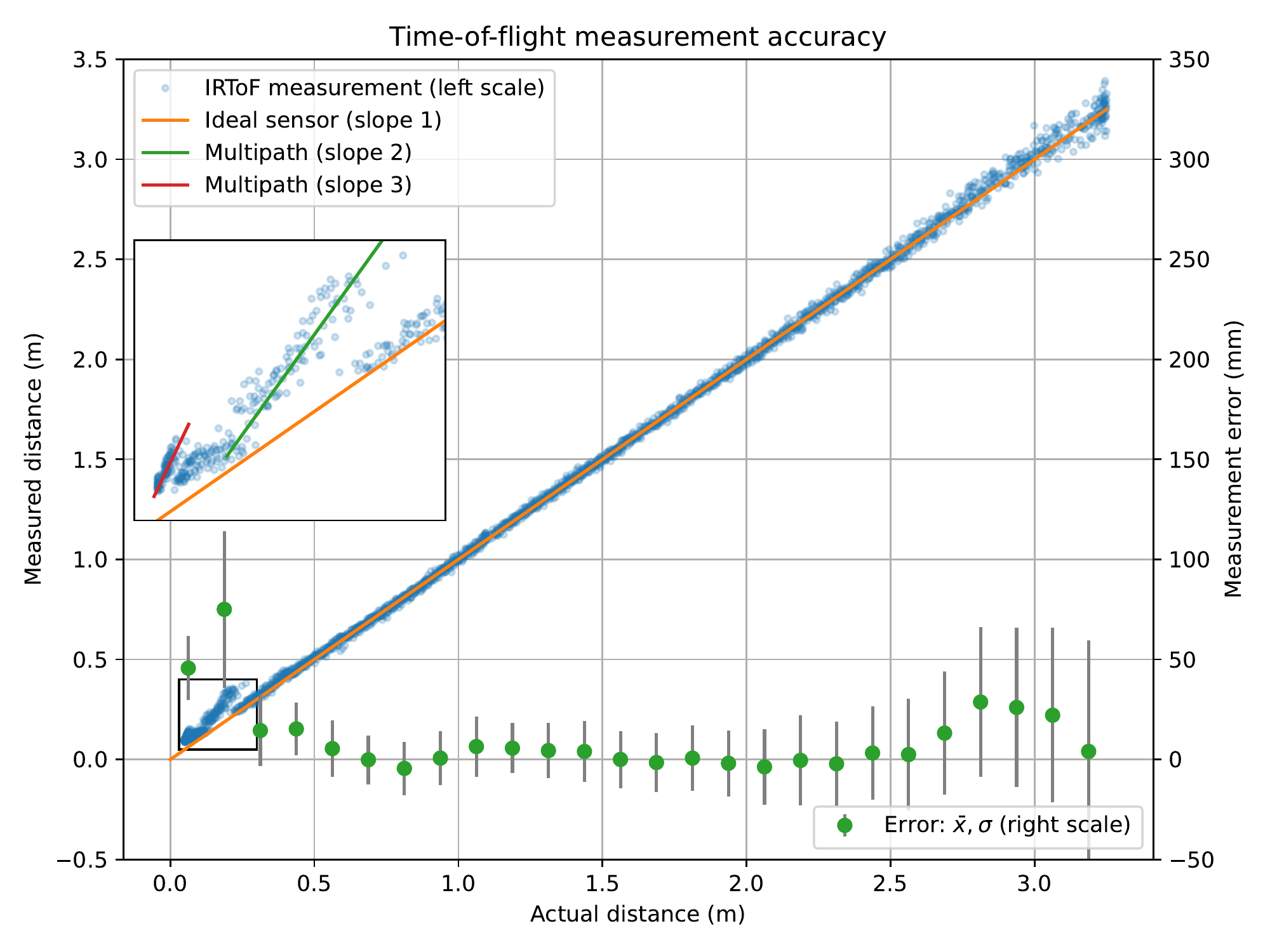}
  	\caption{Plot of 2455 points of IRToF measured distance vs ground truth (left scale) and measurement error in \SI{0.125}{m} bins (right scale). Bars are $\pm \sigma$. The inset shows the data in the range \SI{0.05}{m} to \SI{0.4}{m}. There are signs of multipath effects at close range (less than \SI{0.25}{m}) indicated by the green and red lines of slope 2 and 3.}\label{fig:irtofacc}
\end{figure*}

There is little published data on the performance of the VL53L1CX sensors, outside the manufacturers specifications, so we wanted to characterise this. The sensors were set up according to the manufacturers recommendations using the VL53L1X Ultra Lite driver\footnote{\url{https://www.st.com/en/embedded-software/stsw-img009.html}} for `short' distance mode and update rate of \SI{50}{Hz}. This notionally gives a maximum distance measurement of \SI{1.3}{m}.
We commanded a robot to move repeatedly between locations \SI{0}{m} and \SI{3.5}{m} from a vertical wall made of polypropylene blocks at a speed no greater than \SI{0.5}{ms$^{-1}$} along a path at right angles to the wall. The robot pose was such that one sensor was directly pointing at the wall. The ground truth of the robot position and the distance readings from the IRToF sensor orthogonal to the wall were captured. A total of 2455 measurement samples were collected. Figure \ref{fig:irtofacc} shows the results. Each measured distance is plotted against the ground truth distance of the sensor from the wall. We divided the measurements into bins \SI{0.125}{m} wide and calculated the mean and standard deviation of the error in each bin. 

The performance of the sensor is remarkably good over most of its range. Once above \SI{0.25}{m}, up to \SI{2.7}{m}, the mean measurement error does not exceed \SI{12}{mm} and is mostly less than \SI{5}{mm}. Even out to \SI{3.2}{m}, way beyond the notional range for its mode of operation, the mean error never exceeds \SI{30}{mm}. 

Of some interest is the behaviour of the sensor at close distances, less than \SI{0.25}{m}. Here, the mean error is much higher, up to \SI{75}{mm}, and examining the measured points in more detail (see inset in Figure \ref{fig:irtofacc}) we can see that many of the points do not fall on a line of slope 1, which would be expected from a perfect sensor, but on lines of slope 2, and possibly slope 3. This would be characteristic of multipath reflections e.g. reflecting off the wall, then the robot body, then the wall again before being sensed. 
We intend to investigate other device setup possibilities to see if it possible to mitigate this behaviour.

\paragraph{Additional sensors}
In addition to the cameras and proximity sensors, the robot is equipped with other sensors to understand the robot state or the state of the environment - 9 degree-of-freedom (DoF) Inertial Measurement Unit (IMU), temperature, pressure, humidity, two MEMS microphones, robot health, and auxiliary ADCs for future use. 

The IMU consists of a \SI{6}{DoF} LSM6DSM accelerometer and gyroscope and a \SI{3}{DoF} LIS2MDL magnetometer. The IMU, together with information about wheel position, is the main way the robot performs odometry to estimate its local movement. We currently acquire samples at 100Hz and filter with an EKF to match the motion control update rate. 

The temperature, pressure and humidity sensors allow us to monitor the state of the robots environment - useful in a warehouse scenario, for example, noting locations with excessive humidity or temperatures outside desired limits.

The two IM69D130 MEMS microphones are processed by the on-board FPGA to a I$^2$S stream which is sent to the SBC. The audio is available using the standard Linux audio framework. We intend to use these to experiment with audio-based localisation - sending ultrasonic signals at known times and measuring time-of-arrival to estimate distances.

Robot health sensors monitor the state of various subsystems on board. We measure the voltages and currents of each power supply, the state of each cell of the battery pack, and the temperatures of the battery charging system, power supplies, motor drives, and single board computer. All this information is available through the Linux \textit{hwmon} sensor framework\footnote{\url{https://www.kernel.org/doc/html/latest/hwmon/hwmon-kernel-api.html}}.

\subsubsection{Communication}
Local communication is important for swarm robotics, and an area where there is much potential for novel approaches. To facilitate experimentation, we include multiple different radios. As well as the WiFi built in to the single board computer, each robot is equipped with two nRF52840-based BLE5 (Bluetooth Low Energy) USB-connected radio modules, and a DWM1001 UWB (Ultra wide band) module. A private 5G modem can be added. All of these modules can be reprogrammed with custom firmware under the control of the single board computer, allowing for on-the-fly installation of novel communication protocols.

We currently have firmware for the BLE radios that continually advertises the robots unique name and a small amount of data, and scans for other BLE radios. Each scan result contains the name of any nearby robot sensed, along with its received signal strength (RSSI), which can be used as a proxy measure for distance.  

The DWM1001 radio is designed to perform two-way ranging between devices, measuring the distance between a pair of radios to a claimed accuracy of around 0.1m. It is interfaced to the SBC using an SPI bus.

\subsubsection{Mobility}
For the collective transport of large loads, it is necessary for multiple robots to be able to move together in a fixed geometric relationship with each other. The only way to achieve this with arbitrary trajectories is for the robots to have holonomic motion. We use a three omniwheel system with the wheels equally spaced \SI{120}{\degree} apart. The kinematics for this type of drive are well known \cite{rojas2006holonomic} and are shown below:
\begin{align*}
\begin{pmatrix}
	v_1 \\
	v_2 \\
	v_3 
\end{pmatrix}	=
\begin{pmatrix}
	-\frac{\surd{3}}{2} & 0.5 & 1 \\
	0 & -1 & 1 \\
	\frac{\surd{3}}{2} & -0.5 & 1  
\end{pmatrix}
\begin{pmatrix}
	v_x \\
	v_y \\
	R\omega
\end{pmatrix}
\end{align*}
where $v_1,v_2,v_3$ are the tangential wheel velocities, $v_x,v_y,\omega$ the robot body linear and angular velocities, and $R$ the radius from centre of robot to wheels.

Another important requirement for the robots is that they can move fast and accurately. 
A limitation of other lab-based swarm systems is that the locomotion is often based on stepper motors or small geared DC motors. These are relatively cheap and accurate but are slow and heavy. Much higher performance is possible with Brushless DC (BLDC) servo motors. These servo motors are paired with a position encoder and drive electronics that modulates the coil current to achieve accurate torque, velocity and position control with much higher performance to weight and size ratios than is typically possible with stepper motors. This comes at a cost, a typical motor, encoder, and driver with the performance we require costs around \pounds 350.

\begin{table}\centering\caption{Motor drive cost comparison. Maxxon parts chosen to give broadly similar performance}\label{tab:motorcosts}
{\RaggedRight
\begin{tabulary}{\textwidth}{Lp{0.25\columnwidth}Rp{0.25\columnwidth}R}
	\hline
	Part & Maxxon & Cost & Commodity & Cost \\
	\hline
	Motor & EC45 flat 120W P/N 608148 & 129 & Sunnysky X2814 900Kv & 22 \\
	Encoder &  MILE 2048 P/N 673027 & 97 & Sensor PCB and magnet & 10 \\
	Controller  & ESCON P/N 438725 & 127 & Custom drive electronics & 25 \\
	\hline
	&& \pounds353 & & \pounds57 \\
	\hline
\end{tabulary}
}
\end{table}

There has been recent interest in using commodity drone motors in place of dedicated servo motors \cite{katz2019mini,lee2019empirical,grimminger2020open,kau2019stanford}. The high power-to-weight ratios and low costs due to the large market size make this an interesting alternative. There are disadvantages - the motors are not designed for use as servos and certain parameters which are more important in servo motors are not controlled, e.g. the amount of torque cogging, and the motors often use lower voltages and higher currents than typical servos, but these deficiencies can be compensated for with clever software, e.g. \cite{dini2019cogging}. We designed drive circuitry suitable for running the Odrive Robotics\footnote{\url{https://odriverobotics.com/}} open source BLDC servo controller  and tested various cheap drone motors, selecting for the motor with the least cogging that met the required form factor and price. We replaced the costly traditional optical position encoder with a high resolution magnetic field angle encoder IC. By glueing a small diametrically magnetised disc magnet to the end of a motor shaft and positioning the encoder IC correctly in line with the motor axis, we can sense absolute motor shaft angle with high precision and low cost. 

\begin{figure}
  	\centering
  	\includegraphics[width=1.0\columnwidth]{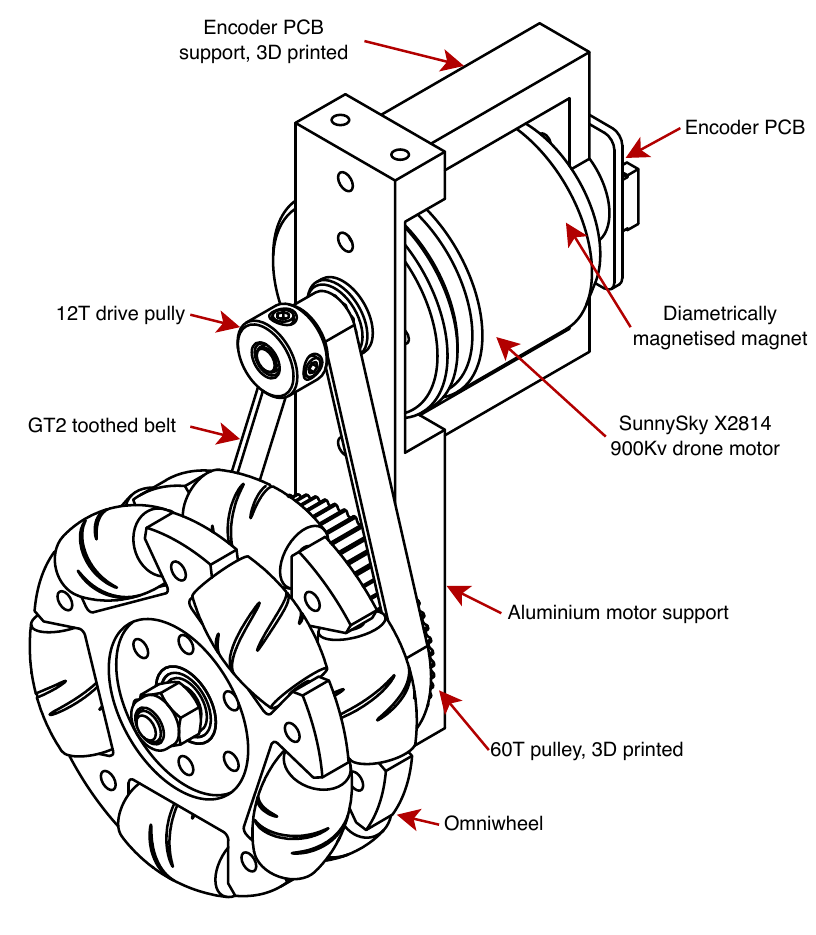}
  	\caption{Single drive assembly, showing commodity drone motor with magnetic angle encoder PCB driving omniwheel via 5:1 toothed belt reduction.}\label{fig:motordrive}
\end{figure}

These innovations reduced the cost per robot of the motor drives from over \pounds 1000 to less than \pounds 200 with comparable performance. Table \ref{tab:motorcosts} illustrates this, and Figure \ref{fig:motordrive} shows a single complete wheel assembly.

\begin{figure}
  	\centering
  	\includegraphics[width=0.98\columnwidth]{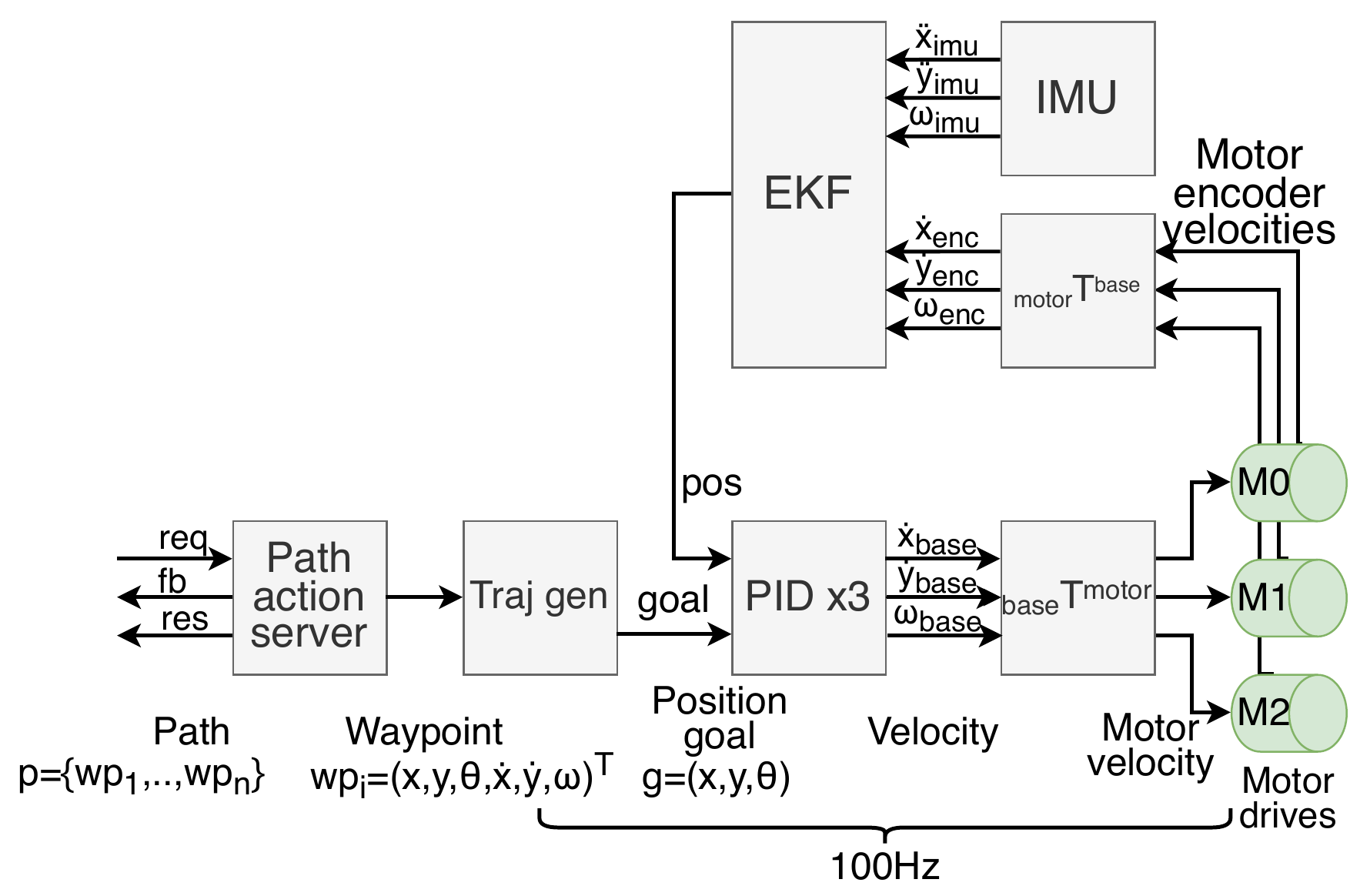}
  	\caption{Motion control system. A ROS action server accepts multi-waypoint paths which get played out by the constrained trajectory generator at \SI{100}{Hz}. The PID control loop operates in the robot local frame, with robot velocities transformed into the appropriate motor velocities.}\label{fig:motion}
\end{figure}

Many systems aimed at swarm robotics research e.g. epucks, pheno, kilobots etc  (cite) are physically quite small and move at low speeds. Collisions are not of high enough energy to cause damage and the mass of the robots is low enough that little attention has to be paid to concerns that are important in larger robots such as trajectory generation respecting acceleration limits. The DOTS robots, however, have a mass of \SI{3}{kg}, with a potential payload of \SI{2}{kg}, so a proper trajectory generation and motion control loop is important. We want to be able to generate new trajectory points in real time with low computational cost, and also to be able to revise the goals on-the-fly. This capability is important for agile quadrotor drones, so there has been recent work in this area. We use the Ruckig trajectory generator described in \cite{berscheid2021jerk}, which is available as C++ library\footnote{\url{https://github.com/pantor/ruckig}}. The on-board motion control system is shown in Figure \ref{fig:motion}. The action server can accept a path consisting of multiple waypoints with required positions and velocities, this is played out by the trajectory generator producing intermediate goal positions at a rate of \SI{100}{Hz}. Three PID controllers  generate velocity commands in the 
 robot frame to satisfy the current goal on the path. Position feedback uses the Inertial Measurement Unit (IMU) and wheel odometry fed through an EKF filter. At any point, the action server can accept a new trajectory or the cancellation of the existing trajectory and goals are updated on-the-fly while respecting acceleration and jerk limits. Cancellation of a trajectory results in maximum acceleration braking to zero velocity. Supplementary material \ref{} shows various examples of agile trajectory following.

\subsubsection{Power}
The power system consists of two Toshiba SCiB 23Ah Lithium Titanate cells, with a total energy capacity of 100 Wh. These are combined into a battery pack, with a battery management system to provide under- and overvoltage protection and monitoring. This is converted by the PSU module to the required voltages \SI{3.3}{V}@\SI{2}{A}, \SI{5}{V}@\si{4}{A}, and \SI{12}{V}@\SI{4}{A}. All voltages, currents, and the battery state are monitored with INA219 power sensors.

A battery charger circuit appropriate to the chemistry of the SCiB cells is integrated into the PSU module, enabling the robot battery to be recharged with a commodity AC-DC adaptor capable of providing \SI{12}{V}@\SI{5}{A}.

Table \ref{tab:poweruse} shows the power consumption of various subsystems. The baseline idle consumption gives access to all sensors except for the vision system, and includes WiFi and Bluetooth communication. At this level, the endurance of a robot with a full charge is around 16 hours. We can see that the consumption increases considerably when using the camera vision system. The motor system figure is for all wheels delivering maximum torque so would not be a continuous state. Average motor power when performing a random walk with a maximum speed of \SI{0.5}{ms$^{-1}$} was around \SI{2}{W}. With full camera usage and moderate movement and processing we get an endurance of around 6 hours.
\begin{table}\centering\caption{Subsystem power consumption. The table indicates the power consumption of some subsystems and battery life under various conditions.}\label{tab:poweruse}
{\RaggedRight
\begin{tabulary}{\textwidth}{p{0.19\columnwidth}ccp{0.4\columnwidth}}
	\hline
	Subsystem & Power (W) & Life$^1$ (h) & Notes\\
	\hline
	Mainboard & 2.5 && Includes all sensors except vision\\
	RockPi4 SBC & 2.5 && Basic telemetry and WiFi networking\\
	 & 9.5 && All 6 CPUs and GPU busy\\
	Vision system & 7.0  && 5x cameras streaming MJPEG 640x480 @30fps\\
	\hline
	Motor system & 0   && Off\\
	& 2 && Random walk \SI{0.5}{ms$^{-1}$} (avg)\\
	& 15   && Full torque on all wheels\\
	\hline
	Total & 6 &14& Basic sensing and comms, occasional movement \\
	& 14 &6& Sensing, comm, and cameras, moderate vision processing and movement \\
	& 34 &2.5& All systems at maximum power consumption\\
	\hline
\end{tabulary}}
	\footnotesize{$^1$Assuming \SI{100}{Wh} battery capacity @85\% efficiency	}
\end{table}

Battery life is limited to the point the output voltage falls below the undervoltage protection limit. The largest dynamic load component is the motor drives, so we compromise on the maximum allowed acceleration in order to limit voltage drop and extend effective endurance. The available torque implies a maximum loaded acceleration of around \SI{4}{ms$^{-2}$} but we limit this to \SI{2}{ms$^{-2}$}.

\subsubsection{Manipulation}
In order to demonstrate novel algorithms for logistics and collective object transport, the robots need a means of carrying payloads. For this purpose, each robot is equipped with a lifter - a platform on top of the robot that can be raised and lowered by the use of a miniature scissor lift mechanism actuated by a high-powered model servo. In the centre of the lifter platform there is an upwards-facing camera that can be used to visually navigate.

To hold payloads, there are carriers, square trays with four legs that the robot can manoeuvre underneath before raising the lifter. In order that a robot can position itself correctly to lift the carrier, we place a visual target on the underside.

Each carrier weighs about \SI{200}{g}, so with the lifting capacity of \SI{2}{kg} we can take a payload of around \si{1.8}{kg}. In order to demonstrate collective transport, the carriers can be joined together along their edges to form arbitrary shapes, to be moved with an equal number of robots, one under each carrier.

\subsection{Industrial Swarm Testbed}
The industrial swarm testbed is the collection of DOTS robots, the physical arena - a \SI{5}{m} x \SI{5}{m} main room overlooked by an adjacent control room, the communications infrastructure comprising WiFi and private 5G network, motion tracking system, video cameras, and the system software architecture. 

It also has an associated high fidelity simulation system, based on the widely used \textit{Gazebo} simulator. Controllers can be safely developed in simulation and transferred without modification to run on the real robots.

This section details the individual components making up the testbed, before tying them together with a description of the complete system architecture.

\subsubsection{Robot Operating System (ROS)}
We use ROS2 Galactic \cite{thomas2014nextgenros} for the testbed. ROS2 is the next generation of ROS \cite{quigley2009ros}, and, at first glance, would appear to have several advantages over ROS1  for the type of decentralised swarm testbed we describe here. Firstly, there is no ROS \textit{master}, ROS2 is completely decentralised. With multiple, potentially communicating, robots, we don't need to choose one to be the master, nor do we need to use multimaster techniques (e.g, \cite{juan2015multi})  as were necessary with ROS1. This is facilitated by the second major difference - the communications fabric of ROS2 uses middleware conforming to the DDS standard \cite{pardo2003omg} of which there are several suppliers. DDS supports automatic discovery of the ROS graph and supports a rich set of Quality of Service (QoS) policies that can be applied to different ROS topics, or communication channels. This maps onto a swarm perspective quite naturally; communications within the ROS graph of a single robot may be marked as \textit{reliable} meaning data will always be delivered, even if retries are necessary, whereas inter-robot communications could be marked as \textit{best effort} meaning data may be lost and dropped with no errors or retries if the physical channel is unreliable. 

However, this enticing vision has yet to be borne out in reality. The DDS discovery process generates network traffic that increases approximately $O(n)$ where $n$ is the number of participants when using UDP multicast, or $O(n^2)$ when using unicast \cite{an2014content} \cite{sanchez2011bloom}. A participant is typically a ROS2 process of which there might be many per robot. Complete information is maintained for all participants at every location, even if they will never communicate. Additionally, although the standard discovery protocol uses UDP multicast, this is very unreliable on WiFi networks \cite{kumari2021rfc} forcing the use of unicast discovery. Building robot systems with many mobile agents linked using wireless communication is becoming more common, and the limitations of ROS2 in this regard have resulted in several possible solutions. One approach, the discovery server, uses a central server to hold information about all participants, greatly reducing traffic. This so far is limited to a single DDS vendor, eProsima\footnote{\url{https://www.eprosima.com/}}, and is the opposite of the decentralised vision. Another approach uses bridges between DDS and a more suitable protocol, this is the approach we used; Zenoh\footnote{\url{https://zenoh.io/}} is an open-source project supported Eclipse Foundation\footnote{\url{https://www.eclipse.org/}}. The Zenoh approach is agnostic of the particular DDS middleware used, and is intended to be part of the next release of ROS2. The Zenoh protocol is designed to overcome the deficiencies of DDS regarding discovery scalability \cite{zenoh2021discovery}. By using a Zenoh-DDS bridge on each agent (robots and other participants) and disallowing DDS traffic off robot, we can achieve transparent ROS2 connectivity with far lower discovery traffic levels, granular control over the topics that can seen outside the robot, true decentralisation, and no use of problematic multicast.

A number of ROS nodes are always running on each robot, interfacing to the hardware and providing a set of ROS topics that constitute the robot API. This API is available on both the simulated and real robot, and is intended to be the only way a controller has access to the hardware.

\subsubsection{Coordinate frames}
We adhere to the ROS standard for coordinate frames REP-105 \cite{ros2010rep105}, see the arena diagram Figure \ref{fig:frames}, with the extension that each frame name that is related to an individual robot is prefixed with the unique robot name. The global frame \textit{map} is fixed to the centre of the arena, \textit{odom} is fixed to the starting position of the robot, and \textit{base\_link} is attached to the robot body, with $+x$ pointing forward, between the overlapping cameras. The \textit{base\_link} frames are updated relative to the \textit{odom} frames using each robots inertial measurements and wheel velocities. There is no transform between \textit{map} and any \textit{odom} unless some form of localisation is used on a robot, which is not necessary for many swarm applications. The ROS transform library TF \cite{foote2013tf} is used to manage transforms. Each robot has its own transform tree, and a transform relay node is used to broadcast a subset of the local transforms to global topics for use in visualisation and data gathering.

\subsubsection{Containerisation}
Controllers for the robots are deployed as Docker \cite{merkel2014docker} containers. Docker provides a lightweight virtualised environment making it easy to package an application and associated dependencies into a container, which can then be deployed to computation resources using an orchestration framework such as Kubernetes, we use the lightweight implementation K3S\footnote{\url{https://k3s.io/}}.  

Containers only have controlled access to the resources of the compute node, facilitating security and abstraction - as far as the container is concerned, it has no knowledge of whether it has been deployed onto a real robot, or a compute node communicating with a simulated environment.

\subsubsection{Real world}
The arena consists of a \SI{5}{m} x \SI{5}{m} main room, overlooked by an adjacent control room. The main room is equipped with an OptiTrack \cite{optitrack2022mocap} motion capture system with 8x Flex 13 120Hz cameras for analysis of experiments. There are separate high resolution video cameras. Communications with the robots are provided with a dedicated \SI{5}{GHz} WiFi access point and a private 5G base station. A Linux server is used for data collection, Kubernetes orchestration, video streaming, and general experiment management. The arena server and 5G base station are connected to the UMBRELLA network \cite{farnham2021umbrella} with a fibre link.

\subsubsection{Simulation}
The standard robot simulator \textit{Gazebo} is used. The simulation environment consists of a configurable model of the real arena with various props such as carriers the robots can pick up, and obstacles, a Unified Robot Description Format (URDF) model of the robot and some of its senses, and a ROS node that provides the same topic API that is present in the real robots.

The robot is modelled taking into account the trade-off between speed and fidelity. Rather than modelling the physically complex holonomic drive omniwheels we instead modelled the motion of the robot as a disc with friction sliding over the floor, with a custom plugin using a PID loop to apply force and torque to the robot body in order meet velocity goals. This has the advantage over directly applying velocity of avoiding unphysical effects such as infinite acceleration, and results in realistic behaviour at low computational cost.


The cameras, time-of-flight proximity sensors, and IMU (Inertial Measurement Unit) are modelled using standard Gazebo ROS plugins and the ROS node emulates hardware sensors such as the battery state and controls actuation of the simulated lifting platform, presenting the same interface as the real hardware.

Simulation is not reality - there are always differences. The \textit{reality-gap} \cite{mouret201720} means that keeping in mind the limitations of simulation when designing robot behaviours that will transfer well to real robots is important. For example, collisions are hard to simulate, and bad for robots, so behaviours relying on them in simulation should be avoided.

\subsubsection{Online portal for remote experimentation}
The physical arena and the simulation environment are designed to be useable remotely. The UMBRELLA Platform, detailed in \cite{farnham2021umbrella} provides cloud infrastructure to facilitate this. An online portal is used for managing experiments through which users may upload controller containers and simulator and experiment configuration files - controlling for example the number and starting positions or the robots, and download data generated during an experimental run. 

The user can schedule or queue experiments to be run in simulation or on the real robots. Because the real robots could be damaged by collisions, the controller containers used for experiments on them need to be verified in a simulation run that checks robot trajectories for dangerous approach velocities and other potential indicators of hazard. This is an open area of research \cite{pickem2016safe,beltrame2018engineering,eder2021complete}.

Controller containers for simulation are run on cloud machines of the same processor architecture as the robots (ARM 64 bit), communicating over ROS topics on a virtual network with a server running a Gazebo simulator instance. When the controller container has been verified and is to be run on the real robots, it is passed to the testbed Linux server and queued to be run by the testbed technician, who ensures the physical arena is correctly set up and the right number of robots positioned. 

 As well as data collected by the experiment containers, the testbed server collects ground truth data from the motion capture system and video from the arena cameras to be streamed to the UMBRELLA cloud system and made available to through the online portal.
 
 \subsubsection{System Architecture}
\begin{figure}
  	\centering
  	\includegraphics[width=1.0\columnwidth]{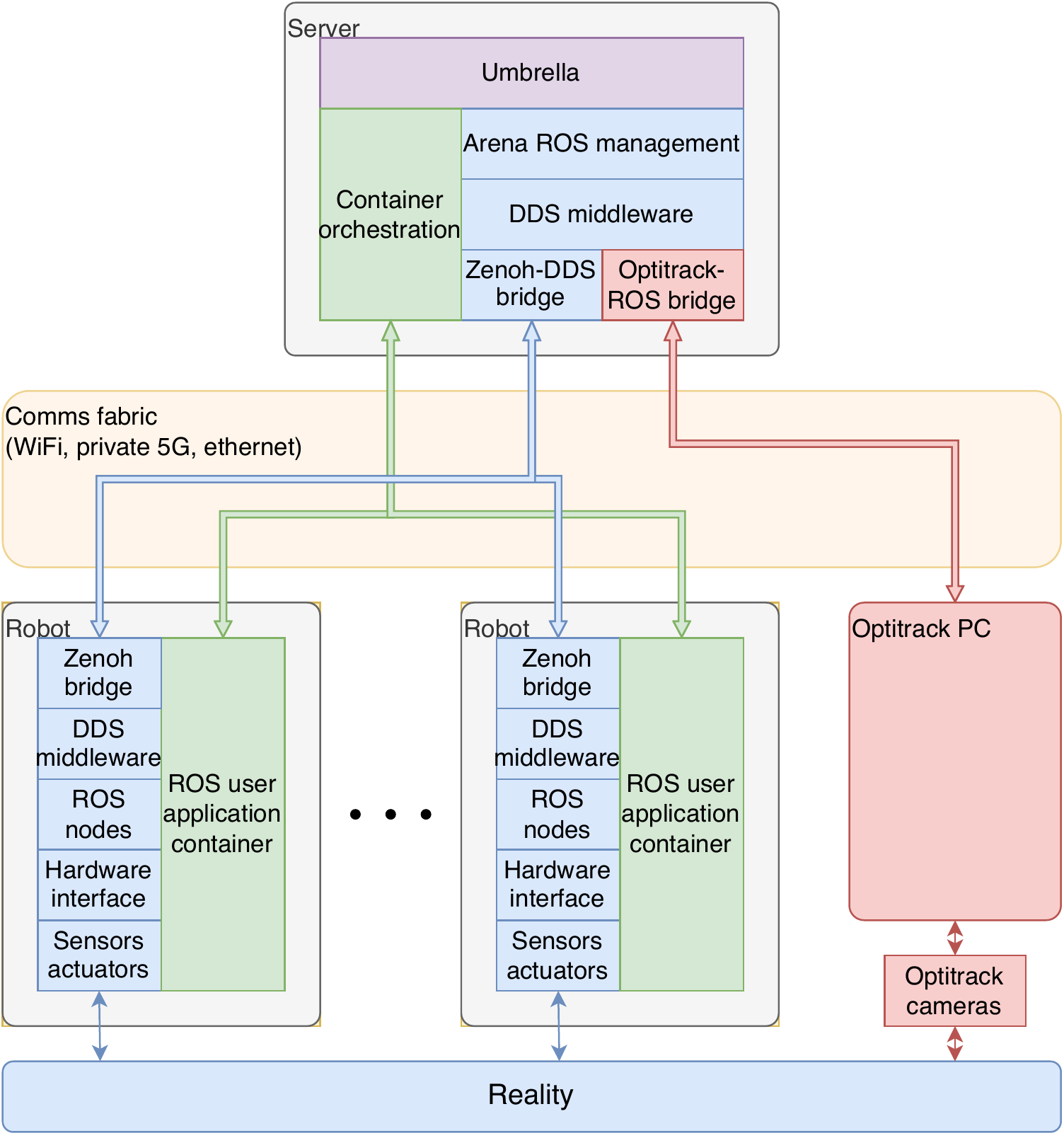}
  	\caption{System architecture. Each robot has a stack of low-level software interfacing to the sensors and actuators, with some ROS nodes running natively and a Zenoh-DDS bridge to handle DDS traffic over the arena communications fabric. Docker containers with ROS user applications are deployed by container orchestration from the server, which also handles Optitrack, cameras, and Umbrella integration. }\label{fig:arena_arch}
\end{figure}

Figure \ref{fig:arena_arch} shows the testbed system architecture. Robots in the arena all run a stack of hardware interface code, housekeeping and low-level ROS nodes that provide the robot API, and communication bridges. User applications or controllers are deployed to the robots as Docker containers using Kubernetes orchestration. A dedicated PC runs an Optitrack motion capture system. A Linux server PC is responsible for running container orchestration, arena system control, camera, ROS, and Optitrack data capture, and interface to the UMBRELLA Platform. 

\subsection{DOTS integrated development environment}
In order to facilitate wider use of the DOTS system, we created a Docker-based development and simulation environment, capable of running on Windows, Linux, and OSX operating systems. There is a large barrier to entry in learning to program  ROS-based robots - versions of ROS are closely tied to particular releases of Ubuntu, dependencies are not always straightforward to install, custom plugins for the simulator need to be compiled, and there are few public examples of complete systems.

Because Docker containers can run on multiple operating systems, we can package up the particular version of Ubuntu, ROS, and all the difficult to install dependencies and provide easy access to a complete ROS2 DOTS code development and simulation environment. All access is provided via web interfaces using open-source technologies developed for cloud applications. The local file system is mounted within the Docker container.

\subsubsection{Code-server}
Code-server\footnote{\url{https://github.com/cdr/code-server}} takes the popular open-source Microsoft editor VSCode\footnote{\url{https://code.visualstudio.com/}} and makes it available via a browser window. We install this within the Docker container. By connecting to a particular port on the localhost, a standard VSCode interface is available in the browser. This not only allows browsing and editing of files, but also provides a terminal for command line access to the environment.

\subsubsection{Gazebo}
The Gazebo simulator is architecturally split into \textit{gzserver} which runs the simulation, and \textit{gzclient} which presents the standard user interface on a Linux desktop. We make available GZWeb, which is another client for \textit{gzserver} that presents a browser-based interface using WebGL \cite{khronos2011webgl}. This allows browser access and is more performant than using the standard interface via VNC.

\subsubsection{VNC}
VNC is a technology for allowing remote desktop access. We run a virtual framebuffer within the Docker container together with a simple window manager and a standard VNC server, and use the noVNC\footnote{\url{https://novnc.com/info.html}} client to make this available through a browser window. This gives access to a standard Linux desktop and allows the use of other graphical applications such as the ROS data visualiser \textit{rviz}.

These three applications packaged in a Docker container enable rapid access to a complete DOTS development and simulation environment in a platform agnostic way. Once a controller application has been prototyped, it can be converted into a Docker container for upload to the UMBRELLA online portal to be verified and for deployment to the real robots.


%

\section{Results}\label{sec:demo}

\subsection{Intralogistics use case}
In order to demonstrate the whole system, we implemented a conceptually simple swarm intralogistics task. Imagine a cloakroom, similar to the scenario described in \cite{jones2020distributed}, where users can deposit and collect bags and jackets, and a swarm of robots will move these to and from a storage area. It is possible to perform this search and retrieval task in an entirely distributed, decentralised manner. Here we implement one aspect of the task, item retrieval, in this decentralised fashion.

The arena, shown in Figure \ref{fig:foragetask}, has two regions, the search zone on the left-hand side, everywhere with $x<0$, and the drop zone on the rightmost 0.6m, everywhere with $x>1.25$. Robots start randomly placed in the drop zone, and carriers in the search zone. The task is for the swarm to find and move the carriers into the drop zone. Apart from the earlier described sensors, we make available two synthetic senses: a compass that gives the absolute orientation of a robot, and a zone sense, indicating if the robot is in either of the two zones. Currently these are derived from the visual localisation described in Section \ref{sec:visloc} but could use, for example, the magnetometer and Bluetooth beacons.

\begin{figure}
  	\centering
  	\includegraphics[width=0.8\columnwidth]{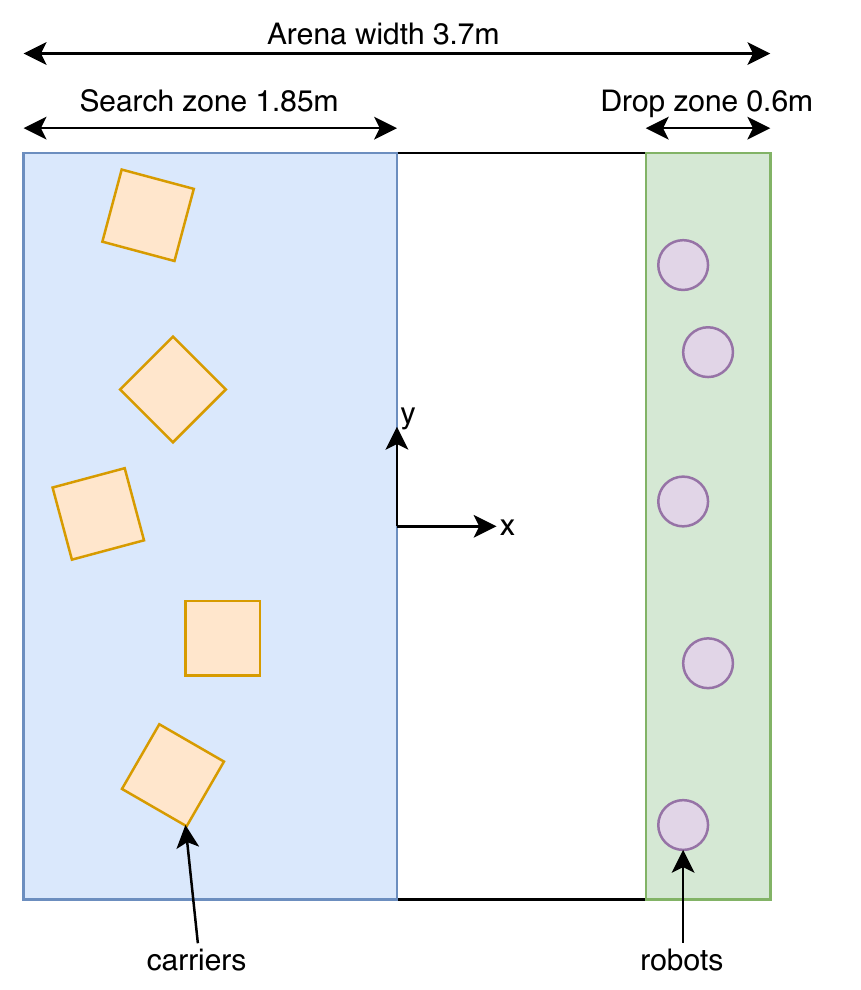}
  	\caption{Intralogistics task, robots starting on the right must search for and pick up carriers in the search zone on the left. Once picked up, the carriers must be moved to the drop zone and deposited.}\label{fig:foragetask}
\end{figure}

Although conceptually simple, for a robot to detect, manoeuvre underneath a carrier and position itself correctly in the centre, then lift it, is non-trivial. Many swarm foraging tasks demonstrated with real robots have used virtual objects \cite{pitonakova2018information}, objects that can be moved just by collision and pushing \cite{jones2019onboard}, or have an actuator that mates with a particular structure \cite{dorigo2013swarmanoid}. These approaches abstract the difficulty of an arbitrary object collection task to various extents. We argue the system we demonstrate here is closer to real-world practical use because the location and lifting of the carriers require the sorts of sensory processing and navigation typical of many non-trivial robotics tasks. 

We develop using the pipeline shown in Figure \ref{fig:pipeline}. The DOTS integrated development environment is used to write controller code, in a mixture of Python and C++, which can be quickly iterated to run on the local Gazebo simulator. Simple sub-behaviours are built up into the complete controller until multiple robots are running successfully in simulation. Once satisfied with simulation the controllers are packaged into Docker containers for validation on the remote UMBRELLA cloud service portal. This allows both larger simulations, with resources limited only by cloud availability, and state-of-the-art validation of the simulated controller for safety. Once validated, the controller may be deployed onto the real robots in the testbed for an experiment run. During the run, multiple data sources are captured and these are then made available for download from the portal.

\subsection{Robot behaviour}
\begin{figure}
  	\centering
  	\includegraphics[width=0.7\columnwidth]{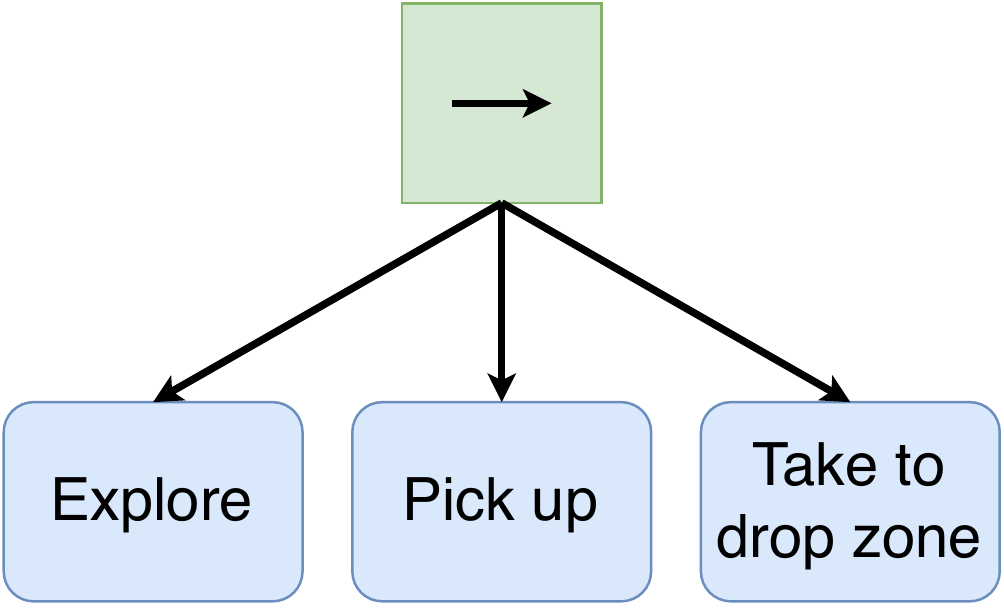}
  	\caption{Top level of Behaviour Tree controller. Robots \textit{explore} until they find a carrier, the \textit{pick up} the carrier, then \textit{take to drop zone} the carrier.}\label{fig:bttask}
\end{figure}

We use a Behaviour Tree controller \cite{champandard2007behavior, ogren2012increasing, klockner2013behavior, jones2016evolving, colledanchise2017behavior}. The top level is shown in Figure \ref{fig:bttask}. There are a sequence of actions that are performed, once each action has been performed successfully, the next is started. Firstly, the robot \textit{explores} the environment, searching for a cargo carrier. If one is found, the \textit{pick up} behaviour is started. This involves manoeuvring under the carrier, positioning itself centrally, then raising the lifting platform. Finally, the robot does \textit{take to drop zone}, moving in the direction of the nest region and once there, lowering the lifting platform. If there are any failures, or once a robot has succeeded in all tasks, the behaviour reverts to \textit{explore}.

\subsubsection{Behaviour trees}
Behaviour Trees are hierarchical structures that combine leaf nodes that interact with the environment into subtrees of progressively more complex behaviours using various composition nodes such as \textit{sequence} and \textit{selector}, see \cite{marzinotto2014towards} for more information. The whole tree is \textit{ticked} at a regular rate, in this case 10Hz, corresponding to the controller update rate, with nodes returning \textit{success}, \textit{failure}, or \textit{running}. A \textit{blackboard} is used to interface between the tree and the robot, holding conditioned and abstracted senses and means of actuation. 
The three top-level behaviours are described below in more detail.

\begin{figure}
  	\centering
  	\includegraphics[trim=0 0 0 40,width=1.0\columnwidth]{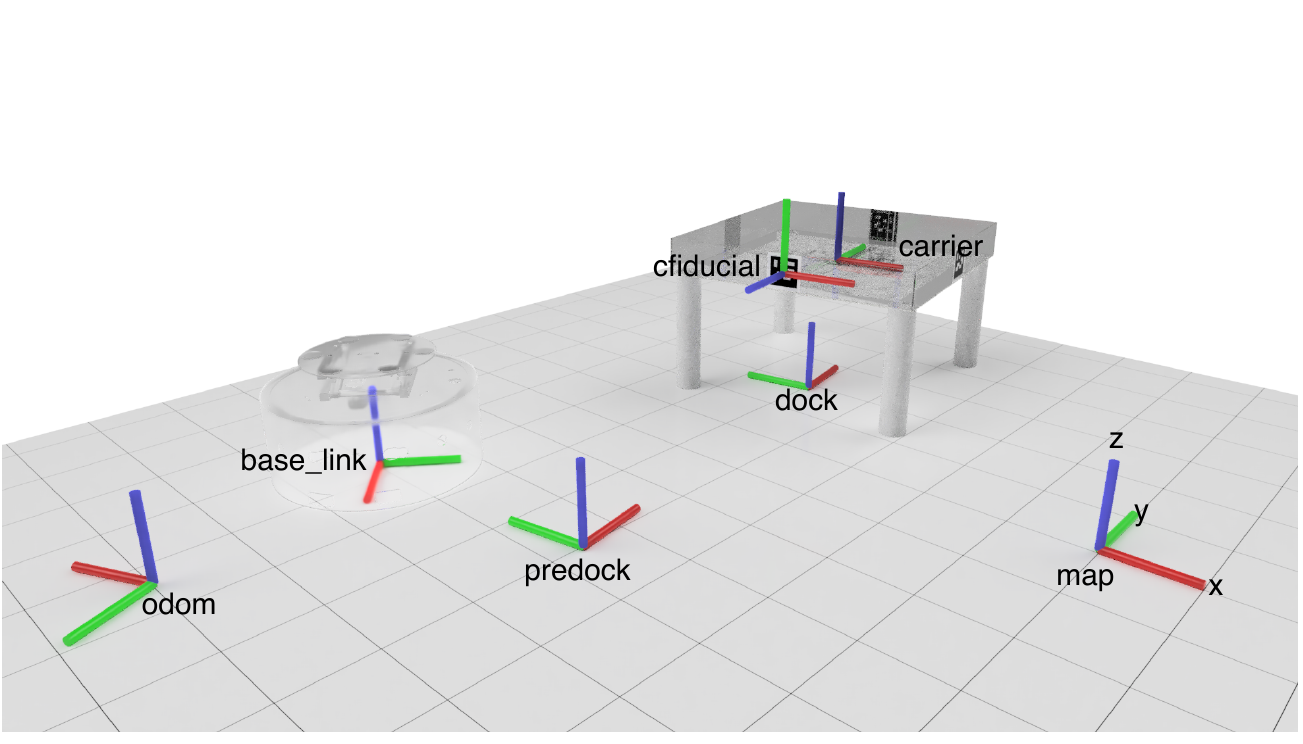}
  	\caption{Frames in use within the testbed. \textit{map} is fixed to the arena, \textit{odom} to the robot starting position. \textit{base\_link} is the robot frame, with $+x$ being forward. Each carrier has four fiducials around the sides, with frame \textit{cfiducial}, and the carrier as a whole has the frame \textit{carrier}, centred in the markermap on its underside. \textit{predock} and \textit{dock} are frames relative to \textit{cfiducial} that the robot uses when navigating under a carrier.}\label{fig:frames}
\end{figure}

\paragraph{Explore}
The \textit{explore} behaviour sends the robot on a ballistic random walk. At the start, or whenever a potential collision is detected, a new direction is chosen and the robot moves in that direction at 0.5ms$^{-1}$. The direction chosen is randomly picked from a Gaussian distribution:
\begin{align}
\theta \sim \mathcal{N}(\pi, \sigma^2) \text{ where }
\begin{cases}
\sigma=3.0\text{ if }	x \in \text{searchzone} \\
\sigma=1.0\text{ otherwise}
\end{cases}
\end{align}
This has probability $p \approx 0.9$ of moving in $-x$ direction when outside the search zone, and $p \approx 0.6$ when in the search zone, so robots will tend to move towards the search zone then randomly within it. The robot continues in the same direction until another potential collision is detected. If, at any point, a carrier ID fiducial (on the four sides), or a carrier markermap (on the underneath) are detected, the robot motion is stopped and the \textit{explore} behaviour  returns \textit{success}, moving to the \textit{pick up} behaviour, otherwise it returns \textit{running}.

\paragraph{Pick up}
This behaviour is started if the robot has seen an ID or markermap. In order of priority, the robot will try to move to the centre position under the carrier if a markermap has been seen, or move underneath the carrier to the \textit{dock} (see frame diagram, Figure \ref{fig:frames}) position if an ID has been seen and robot is close to \textit{predock} position or move to the \textit{predock} position if an ID has been seen. If the centre is reached successfully, the lifting platform is raised and the behaviour returns \textit{success} and starts the \textit{take to drop zone} behaviour. If the vision system loses track of the fiducial ID or the markermap, the behaviour returns \textit{failure}, otherwise it returns running. The consequence of \textit{failure} is reversion to the \textit{explore} behaviour.

\paragraph{Take to drop zone}
This behaviour is started if the \textit{pick up} behaviour has succeeded. It is similar to the \textit{explore} behaviour but biassed to move towards the drop zone. Movement is slower at 0.3ms$^{-1}$ and collision detection suited to the situation where the legs of the carrier can obscure some of the IRToF sensors. This behaviour will return \textit{running} until the robot reaches the drop zone with no collision detection, at which point it will lower the lifting platform and return \textit{success} and revert to \textit{explore} behaviour.

\subsubsection{Cargo carrier detection}
\begin{figure}
  	\centering
  	\includegraphics[width=0.7\columnwidth]{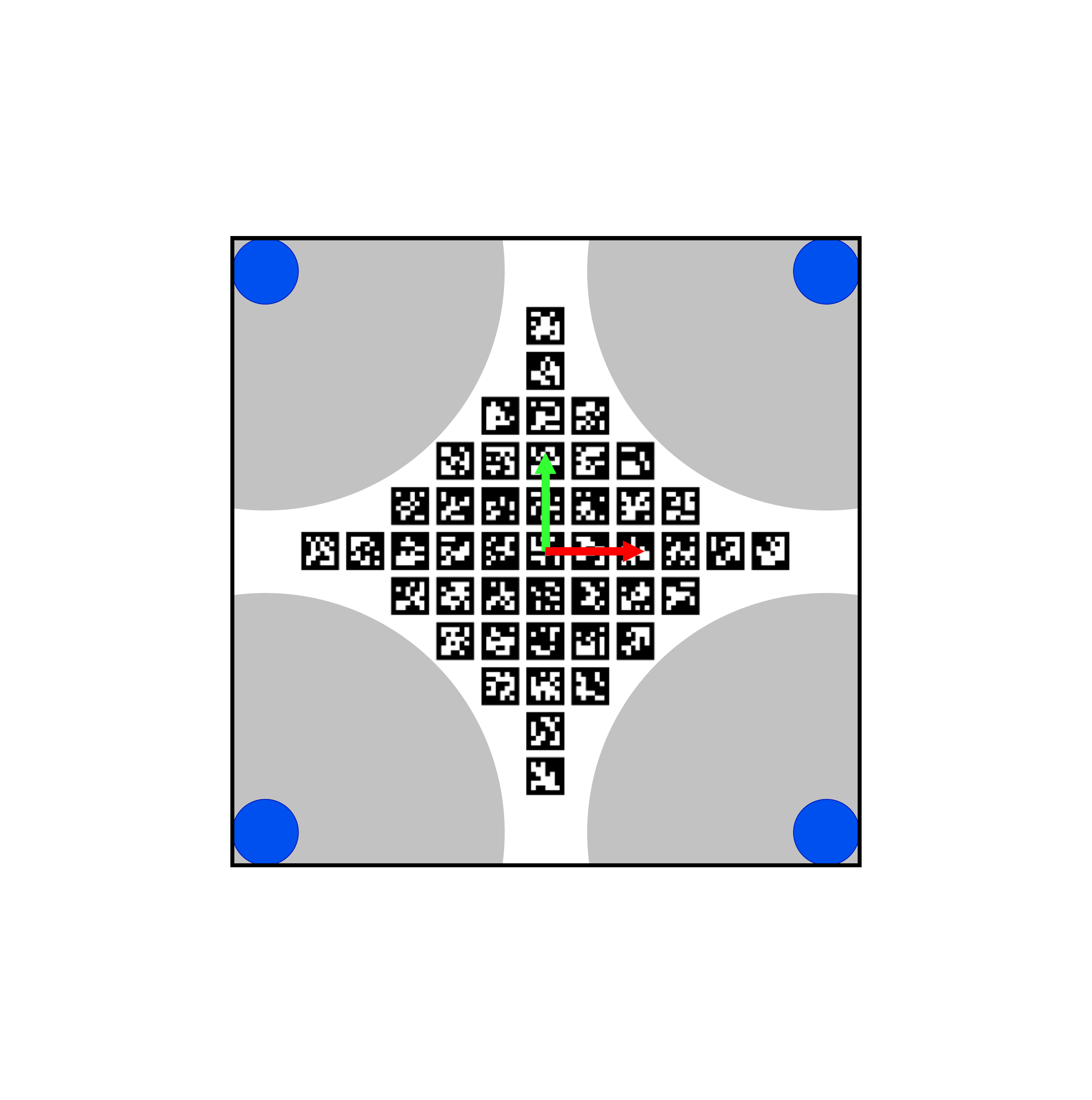}
  	\caption{Underside of a carrier, showing the markermap and the frame. Legs in blue, and the grey areas show regions where the centreline of the upwards-facing camera cannot physically reach.}\label{fig:carriermm}
\end{figure}
Carriers have an ArUco fiducial marker on each side, labelled \textit{cfiducial} in Figure \ref{fig:frames}. These markers has an ID unique to the carrier, and can be detected by the perimeter cameras of the robot from about 1m away. On the underside of the carrier is a markermap, shown in Figure \ref{fig:carriermm}, so the robot can locate the centre of the carrier  using its upwards-facing camera. The size and density of the fiducials making up the map are chosen such that there are always at least one complete marker visible in the camera field-of-view for all physically possible positions of the robot under the carrier. 

When a carrier fiducial has been seen by any camera, the transform from the robot to that fiducial is made available on the \textit{blackboard}. The detection system then focusses on that ID for a short time (3s) and ignores all other IDs, this is to prevent flipping of attention when multiple IDs are visible. The fiducial transform can be used by the controller to navigate to the \textit{predock} and \textit{dock} locations relative to the carrier.

When a carrier markermap is visible, the transform to the centre of the carrier is made available on the \textit{blackboard}. This can be used to navigate the robot to correct central location under the carrier in order to safely lift it.

\subsubsection{Collision}
Information from the IRToF sensors is used to build a dynamic robot-relative map of obstacles. The map is an array of cells, centred on the robot, with values in the range $[0,1]$. The value of a cell continually reduces by exponential decay rate $\lambda$, and is increased when a cell is affected by a sensor return. Sensor returns increase a cell location by a Gaussian based on distance between each return location $(x'_i,y'_i)$ and cell location $(x,y)$:
\begin{align}
	r_i &= \sqrt{(x'_i-x)^2+(y'_i-y)^2} \\ 
	S(x,y) &= \sum_{i=1}^n{e^{-\frac{r_i^2}{2\sigma_{sensor}^2}}}
\end{align}
where $S(x,y)$ is the sensor return function, showing the total effect all $n$ sensor returns have at this location.

Each map location $M(x,y)$ changes over time as:
\begin{align}
M_{t+\Delta t}(x,y) = min(M_t(x,y)(1-\lambda\Delta t) + S(x,y), 1.0)
\end{align}
with the decay rate $\lambda=1$s, and $\sigma_{sensor}=25$mm.
\begin{figure}
  	\centering
  	\includegraphics[width=0.7\columnwidth]{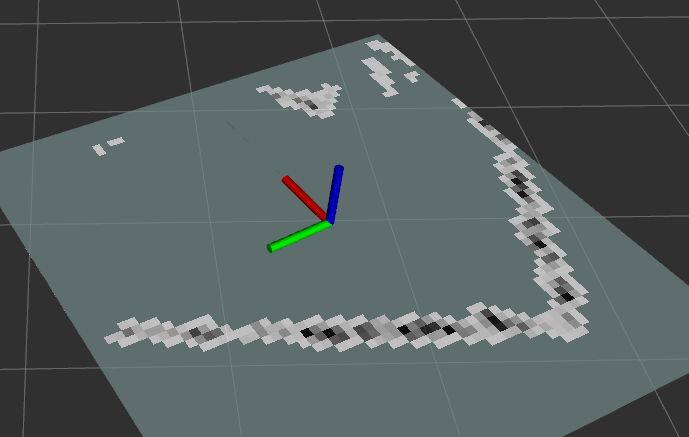}
  	\caption{Example collision map as visualised by \textit{Rviz}, with a frame indicator at the location of the robot, and the grey area the robot-relative map, showing two of the arena walls.}\label{fig:localmap}
\end{figure}

Static objects result in corresponding cells in the map reaching the upper limit of 1.0, transient or false returns tend towards zero.

Collisions are predicted by projecting the current velocity vector forward by a lookahead time, then taking the average value of the cell contents over a circular area related to the robot size. The scalar value is a collision likelihood which is made available on the \textit{blackboard}.

\subsubsection{Navigation}
The various behaviours need to navigate and choose a path to follow. All navigation is performed in the local frame of the robot - there is no global map. For example, when \textit{exploring}, a direction is chosen as described above, then a lower-level navigation behaviour is started that creates a trajectory, then continually monitors the collision map of the locality to abort the trajectory playout if necessary. If there are no paths available that don't trigger a collision warning, for example, if a robot is under a carrier, then a fallback behaviour of picking a least-worst (lowest sum of nearby collision cells) direction and moving slowly for short distances is used to safely emerge into more open space.

%

\begin{figure*}
  	\centering
  	\includegraphics[width=0.85\textwidth]{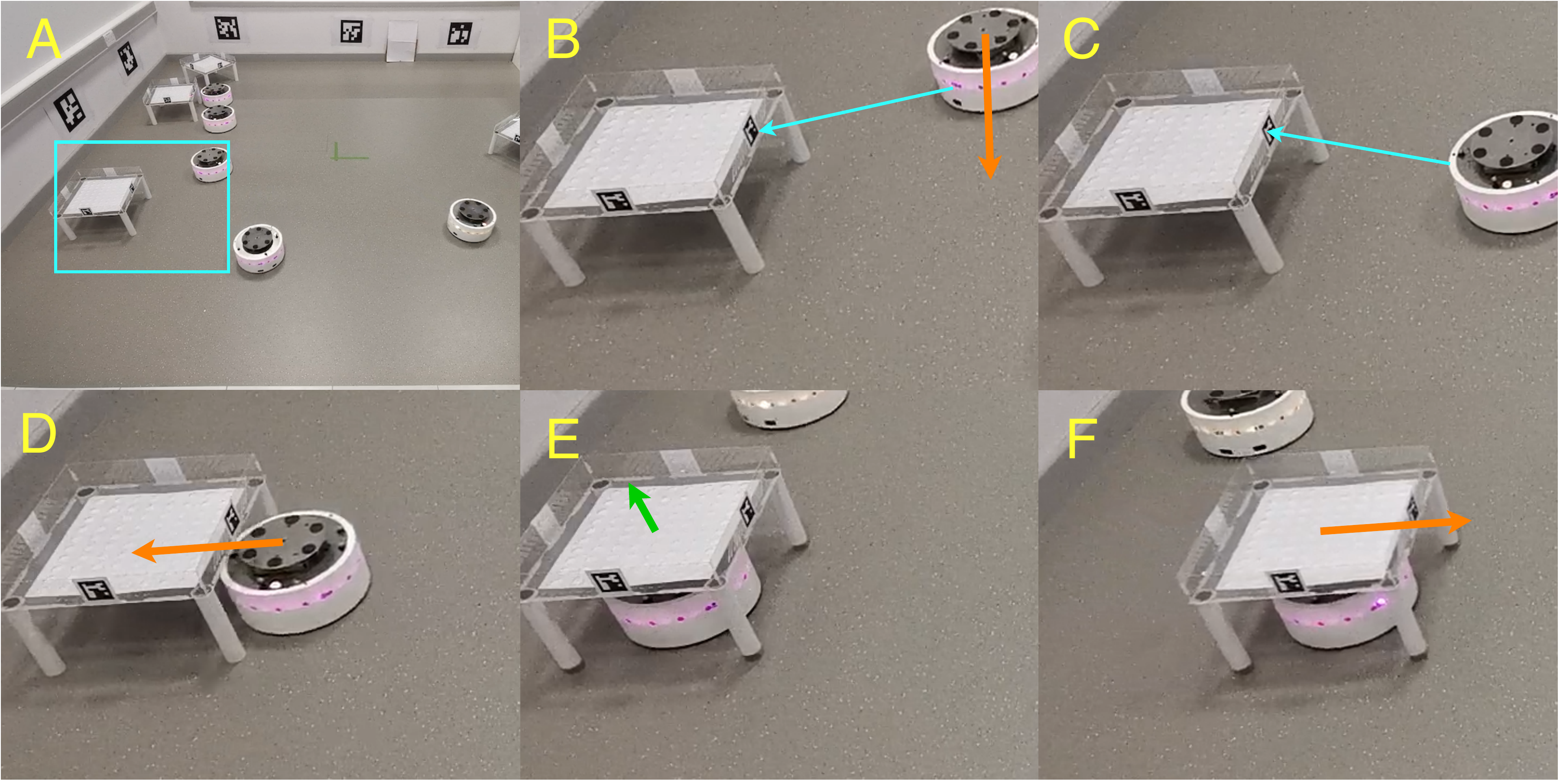}
  	\caption{Sequence showing a robot picking up a carrier. \textit{(A)} Overall View. \textit{(B)} While exploring, a robot detects a carrier fiducial (blue arrow) and starts the \textit{pick up} behaviour. It moves to the pre-dock position (orange arrow). \textit{(C)} The robot is now facing the carrier and is in the optimal position for accurate pose estimation of the carrier fiducial so \textit{(D)} moves under the carrier and \textit{(E)} centres itself using upwards-facing camera and raises the lifting platform. Finally \textit{(F)} the robot moves towards the drop zone.}\label{fig:sequence}
\end{figure*}

\subsection{Performance}
We ran the task five times, in each case the five robots were positioned with random orientation in the drop zone, and the carriers in the search zone. Robots were allowed to run until the task was completed and all robots had escaped the drop zone, or more than 10 minutes elapsed. The minimum time the task could take, assuming each robot goes directly to the nearest carrier, the distance from robot to carrier is \SI{3}{m}, search speed is \SI{0.5}{ms$^{-1}$}, carry speed \SI{0.3}{ms$^{-1}$}, pickup and drop times \SI{5}{s}, is \SI{26}{s}
Obviously, this assumes each robot moves directly underneath a separate carrier, there are no collisions, and they move directly to the drop zone.

\begin{figure}
  	\centering
  	\includegraphics[trim=10 10 10 10,width=1.0\columnwidth]{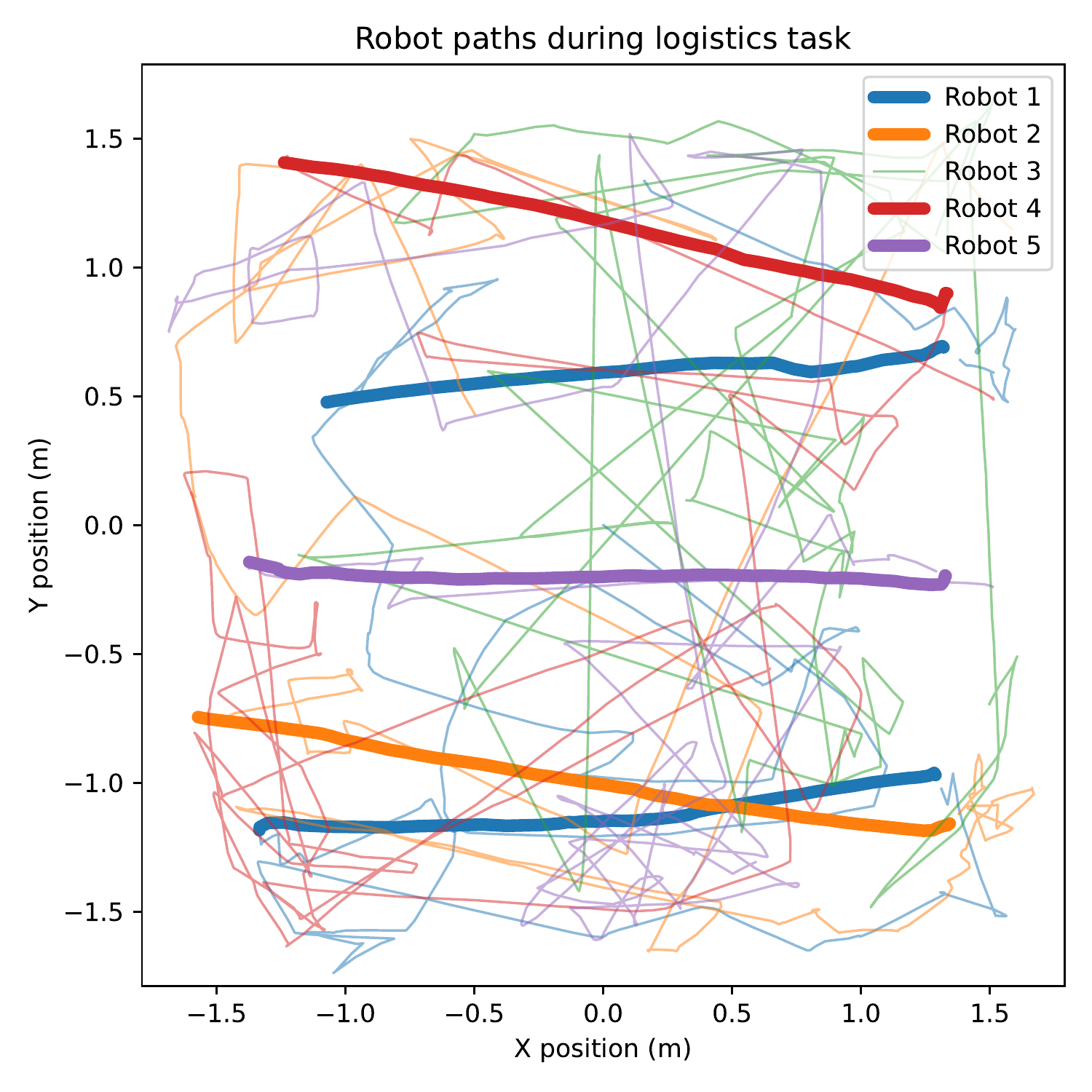}
  	\caption{Tracks of the robots in one run of the logistics task, with thicker tracks when a robot is carrying a load. loads are carried from left to right. In this example, robot 3 never picks up a load.}\label{fig:foragetask}
\end{figure}

Table \ref{tab:tasktime} shows the results. In every case, the swarm was able to successfully retrieve all the carriers within the allowed 10 minutes. The average time to complete the task was 178 seconds. All the runs are shown in the Supplementary video material \href{https://drive.google.com/file/d/1EuA8PS1qpqK6LIfPwCNXtQ3hHNWPDvtN/view?usp=sharing}{logistics task runs}. 

\begin{table}\centering\caption{Performance in intralogistics task}\label{tab:tasktime}
{\RaggedRight
\begin{tabulary}{\textwidth}{CCC}
	\hline
	Run & Carriers retrieved & Time (s) \\
	\hline
	1 &5 & 136\\
	2 &5 & 99\\
	3 &5 & 213\\
	4 &5 & 116\\
	5 &5 & 327\\
	\hline
	&$\bar x$& 178 \\
	\hline
\end{tabulary}
}
\end{table}

\section{Conclusion}\label{sec:concs}
In this work, we have introduced DOTS, our new open access industrial swarm testbed. With a purpose-build arena, 20 high specification robots that move fast, have long endurance and high computational power, a platform-agnostic development environment, and a cloud portal, this system breaks new ground in enabling experimentation.

The intralogistics scenario demonstrates the abilities of the robot swarm to successfully complete a non-trivial task. Although conceptually simple, the underlying processing necessary to handle the five concurrent video streams, time-of-flight pointcloud, and other senses, and entirely autonomously search for, find, manoeuvre under with precision, lift, carry and deposit multiple cargo carriers, is considerable. More importantly, we demonstrate a process. The DOTS Testbed pipeline reduces barriers to entry by making it easy to get started, using the cross-platform IDE to experiment with ideas in simulation, then making available a real physical robot swarm to see these ideas translated into reality. 

There are many directions this research can go in. Currently, there is necessarily some manual management of the robots, they need to be manually charged, set up for experimental runs, and watched over. We are currently designing automatic charging stations to aid this. Validation of controllers in simulation is difficult, although the simulator is relatively high-fidelity, it still suffers from differences to reality, and it is not possible to guarantee the safety of a general-purpose controller; even defining safety for a swarm is an open question. At the moment, we are not utilising the full processing power of the robotic platform - we intend to accelerate image processing, moving it onto the GPUs, both on the single board computer and on the peripheral Raspberry Pi Zeros. We look forward to exploring the possibilities for localisation and short-range communication of the BLE5 and UWE radios, and the two microphones. The possibilities for decentralised stochastic swarm algorithms to reliably perform logistics tasks is touched upon here, and is open to further work. Designing toolkits of useful, working, and validated sub-behaviours from which to more easily construct full applications is another step towards wider real-world use of swarm techniques.

Building swarm robotic applications for the real world is hard, and one of the biggest obstacles is the lack of actually existing robot swarms with the types of capabilities necessary to experiment with ideas. By making available such a swarm, together with an accessible development pipeline, we hope to reduce the barriers to entry and stimulate further research.

\section*{Acknowledgment}

S.J. was supported by EPSRC DTP Doctoral Prize Award EP/T517872/1. S.J. and S.H. were supported by EPSRC IAA Award EP/R511663/1. E.M. was supported by the EPSRC Centre for Doctoral Training in Future Autonomous and Robotic Systems (FARSCOPE) and an EPSRC ICASE Award.
Toshiba Bristol Research and Innovation Laboratory (BRIL) provided additional support.

\ifCLASSOPTIONcaptionsoff
  \newpage
\fi



%

\bibliographystyle{IEEEtran}

\begin{thebibliography}{10}
\providecommand{\url}[1]{#1}
\csname url@samestyle\endcsname
\providecommand{\newblock}{\relax}
\providecommand{\bibinfo}[2]{#2}
\providecommand{\BIBentrySTDinterwordspacing}{\spaceskip=0pt\relax}
\providecommand{\BIBentryALTinterwordstretchfactor}{4}
\providecommand{\BIBentryALTinterwordspacing}{\spaceskip=\fontdimen2\font plus
\BIBentryALTinterwordstretchfactor\fontdimen3\font minus
  \fontdimen4\font\relax}
\providecommand{\BIBforeignlanguage}[2]{{%
\expandafter\ifx\csname l@#1\endcsname\relax
\typeout{** WARNING: IEEEtran.bst: No hyphenation pattern has been}%
\typeout{** loaded for the language `#1'. Using the pattern for}%
\typeout{** the default language instead.}%
\else
\language=\csname l@#1\endcsname
\fi
#2}}
\providecommand{\BIBdecl}{\relax}
\BIBdecl

\bibitem{custodio2020flexible}
L.~Custodio and R.~Machado, ``Flexible automated warehouse: a literature review
  and an innovative framework,'' \emph{The International Journal of Advanced
  Manufacturing Technology}, vol. 106, no.~1, pp. 533--558, 2020.

\bibitem{jaghbeer2020automated}
Y.~Jaghbeer, R.~Hanson, and M.~I. Johansson, ``Automated order picking systems
  and the links between design and performance: a systematic literature
  review,'' \emph{International Journal of Production Research}, vol.~58,
  no.~15, pp. 4489--4505, 2020.

\bibitem{draganjac2016decentralized}
I.~Draganjac, D.~Mikli{\'c}, Z.~Kova{\v{c}}i{\'c}, G.~Vasiljevi{\'c}, and
  S.~Bogdan, ``Decentralized control of multi-agv systems in autonomous
  warehousing applications,'' \emph{IEEE Transactions on Automation Science and
  Engineering}, vol.~13, no.~4, pp. 1433--1447, 2016.

\bibitem{fragapane2021planning}
G.~Fragapane, R.~de~Koster, F.~Sgarbossa, and J.~O. Strandhagen, ``Planning and
  control of autonomous mobile robots for intralogistics: Literature review and
  research agenda,'' \emph{European Journal of Operational Research}, 2021.

\bibitem{csahin2005swarm}
E.~{\c S}ahin, ``Swarm robotics: From sources of inspiration to domains of
  application,'' in \emph{International Workshop on Swarm robotics (SR 2004)},
  S.~W. {\c S}ahin~E., Ed.\hskip 1em plus 0.5em minus 0.4em\relax Santa Monica,
  CA, USA: Springer, 2005, pp. 10--20.

\bibitem{schranz2021swarm}
M.~Schranz, G.~A. Di~Caro, T.~Schmickl, W.~Elmenreich, F.~Arvin,
  A.~{\c{S}}ekercio{\u{g}}lu, and M.~Sende, ``Swarm intelligence and
  cyber-physical systems: concepts, challenges and future trends,'' \emph{Swarm
  and Evolutionary Computation}, vol.~60, p. 100762, 2021.

\bibitem{pickem2017robotarium}
D.~Pickem, P.~Glotfelter, L.~Wang, M.~Mote, A.~Ames, E.~Feron, and
  M.~Egerstedt, ``The robotarium: A remotely accessible swarm robotics research
  testbed,'' in \emph{2017 IEEE International Conference on Robotics and
  Automation (ICRA)}.\hskip 1em plus 0.5em minus 0.4em\relax IEEE, 2017, pp.
  1699--1706.

\bibitem{paull2017duckietown}
L.~Paull, J.~Tani, H.~Ahn, J.~Alonso-Mora, L.~Carlone, M.~Cap, Y.~F. Chen,
  C.~Choi, J.~Dusek, Y.~Fang \emph{et~al.}, ``Duckietown: an open, inexpensive
  and flexible platform for autonomy education and research,'' in \emph{2017
  IEEE International Conference on Robotics and Automation (ICRA)}.\hskip 1em
  plus 0.5em minus 0.4em\relax IEEE, 2017, pp. 1497--1504.

\bibitem{ten2020technical}
M.~ten Hompel, H.~Bayhan, J.~Behling, L.~Benkenstein, J.~Emmerich, G.~Follert,
  M.~Grzenia, C.~Hammermeister, H.~Hasse, D.~Hoening \emph{et~al.}, ``Technical
  report: Loadrunner{\textregistered}, a new platform approach on collaborative
  logistics services,'' \emph{Logistics Journal: nicht referierte
  Ver{\"o}ffentlichungen}, vol. 2020, no.~10, 2020.

\bibitem{williams2009low}
M.~Williams, ``Low pin-count debug interfaces for multi-device systems,'' 2009.

\bibitem{mur2017orb}
R.~Mur-Artal and J.~D. Tard{\'o}s, ``Orb-slam2: An open-source slam system for
  monocular, stereo, and rgb-d cameras,'' \emph{IEEE transactions on robotics},
  vol.~33, no.~5, pp. 1255--1262, 2017.

\bibitem{romero2018speeded}
F.~J. Romero-Ramirez, R.~Mu{\~n}oz-Salinas, and R.~Medina-Carnicer, ``Speeded
  up detection of squared fiducial markers,'' \emph{Image and vision
  Computing}, vol.~76, pp. 38--47, 2018.

\bibitem{bachhuber2017minimization}
C.~Bachhuber, E.~Steinbach, M.~Freundl, and M.~Reisslein, ``On the minimization
  of glass-to-glass and glass-to-algorithm delay in video communication,''
  \emph{IEEE Transactions on Multimedia}, vol.~20, no.~1, pp. 238--252, 2017.

\bibitem{bachhuber2017today}
C.~Bachhuber and E.~Steinbach, ``Are today's video communication solutions
  ready for the tactile internet?'' in \emph{2017 IEEE Wireless Communications
  and Networking Conference Workshops (WCNCW)}.\hskip 1em plus 0.5em minus
  0.4em\relax IEEE, 2017, pp. 1--6.

\bibitem{moore2014ageneralized}
T.~Moore and D.~Stouch, ``A generalized extended kalman filter implementation
  for the robot operating system,'' in \emph{Proceedings of the 13th
  International Conference on Intelligent Autonomous Systems (IAS-13)}.\hskip
  1em plus 0.5em minus 0.4em\relax Springer, July 2014.

\bibitem{su2016material}
S.~Su, F.~Heide, R.~Swanson, J.~Klein, C.~Callenberg, M.~Hullin, and
  W.~Heidrich, ``Material classification using raw time-of-flight
  measurements,'' in \emph{Proceedings of the IEEE Conference on Computer
  Vision and Pattern Recognition}, 2016, pp. 3503--3511.

\bibitem{callenberg2021low}
C.~Callenberg, Z.~Shi, F.~Heide, and M.~B. Hullin, ``Low-cost spad sensing for
  non-line-of-sight tracking, material classification and depth imaging,''
  \emph{ACM Transactions on Graphics (TOG)}, vol.~40, no.~4, pp. 1--12, 2021.

\bibitem{rojas2006holonomic}
R.~Rojas and A.~G. F{\"o}rster, ``Holonomic control of a robot with an
  omnidirectional drive,'' \emph{KI-K{\"u}nstliche Intelligenz}, vol.~20,
  no.~2, pp. 12--17, 2006.

\bibitem{katz2019mini}
B.~Katz, J.~Di~Carlo, and S.~Kim, ``Mini cheetah: A platform for pushing the
  limits of dynamic quadruped control,'' in \emph{2019 international conference
  on robotics and automation (ICRA)}.\hskip 1em plus 0.5em minus 0.4em\relax
  IEEE, 2019, pp. 6295--6301.

\bibitem{lee2019empirical}
U.~H. Lee, C.-W. Pan, and E.~J. Rouse, ``Empirical characterization of a
  high-performance exterior-rotor type brushless dc motor and drive,'' in
  \emph{2019 IEEE/RSJ International Conference on Intelligent Robots and
  Systems (IROS)}.\hskip 1em plus 0.5em minus 0.4em\relax IEEE, 2019, pp.
  8018--8025.

\bibitem{grimminger2020open}
F.~Grimminger, A.~Meduri, M.~Khadiv, J.~Viereck, M.~W{\"u}thrich, M.~Naveau,
  V.~Berenz, S.~Heim, F.~Widmaier, T.~Flayols \emph{et~al.}, ``An open
  torque-controlled modular robot architecture for legged locomotion
  research,'' \emph{IEEE Robotics and Automation Letters}, vol.~5, no.~2, pp.
  3650--3657, 2020.

\bibitem{kau2019stanford}
N.~Kau, A.~Schultz, N.~Ferrante, and P.~Slade, ``Stanford doggo: An
  open-source, quasi-direct-drive quadruped,'' in \emph{2019 International
  conference on robotics and automation (ICRA)}.\hskip 1em plus 0.5em minus
  0.4em\relax IEEE, 2019, pp. 6309--6315.

\bibitem{dini2019cogging}
P.~Dini and S.~Saponara, ``Cogging torque reduction in brushless motors by a
  nonlinear control technique,'' \emph{Energies}, vol.~12, no.~11, p. 2224,
  2019.

\bibitem{berscheid2021jerk}
L.~Berscheid and T.~Kr{\"o}ger, ``Jerk-limited real-time trajectory generation
  with arbitrary target states,'' \emph{arXiv preprint arXiv:2105.04830}, 2021.

\bibitem{thomas2014nextgenros}
\BIBentryALTinterwordspacing
D.~Thomas, W.~Woodall, and E.~Fernandez, ``{Next-generation ROS: Building on
  DDS},'' in \emph{ROSCon Chicago 2014}.\hskip 1em plus 0.5em minus 0.4em\relax
  Mountain View, CA: Open Robotics, sep 2014. [Online]. Available:
  \url{https://vimeo.com/106992622}
\BIBentrySTDinterwordspacing

\bibitem{quigley2009ros}
M.~Quigley, K.~Conley, B.~Gerkey, J.~Faust, T.~Foote, J.~Leibs, R.~Wheeler, and
  A.~Y. Ng, ``{ROS}: an open-source {Robot} {Operating} {System},'' in
  \emph{IEEE International Conference on Robotics and Automation (ICRA 2009)
  Workshop on Open Source Robotics}, vol.~3, no. 3.2.\hskip 1em plus 0.5em
  minus 0.4em\relax Kobe, Japan: IEEE, 2009, p.~5.

\bibitem{juan2015multi}
S.~H. Juan and F.~H. Cotarelo, ``Multi-master ros systems,'' \emph{Institut de
  Robotics and Industrial Informatics}, pp. 1--18, 2015.

\bibitem{pardo2003omg}
G.~Pardo-Castellote, ``Omg data-distribution service: Architectural overview,''
  in \emph{23rd International Conference on Distributed Computing Systems
  Workshops, 2003. Proceedings.}\hskip 1em plus 0.5em minus 0.4em\relax IEEE,
  2003, pp. 200--206.

\bibitem{an2014content}
K.~An, A.~Gokhale, D.~Schmidt, S.~Tambe, P.~Pazandak, and G.~Pardo-Castellote,
  ``Content-based filtering discovery protocol (cfdp) scalable and efficient
  omg dds discovery protocol,'' in \emph{Proceedings of the 8th ACM
  International Conference on Distributed Event-Based Systems}, 2014, pp.
  130--141.

\bibitem{sanchez2011bloom}
J.~Sanchez-Monedero, J.~Povedano-Molina, J.~M. Lopez-Vega, and J.~M.
  Lopez-Soler, ``Bloom filter-based discovery protocol for dds middleware,''
  \emph{Journal of Parallel and Distributed Computing}, vol.~71, no.~10, pp.
  1305--1317, 2011.

\bibitem{kumari2021rfc}
W.~Kumari, C.~E. Perkins, M.~McBride, D.~Stanley, and J.~C. Z{\'u}{\~n}iga,
  ``Rfc 9119-multicast considerations over ieee 802 wireless media,'' 2021.

\bibitem{zenoh2021discovery}
\BIBentryALTinterwordspacing
{Eclipse Foundation}. (2021) Minimizing discovery overhead in {ROS2}. [Online].
  Available: \url{https://zenoh.io/blog/2021-03-23-discovery/}
\BIBentrySTDinterwordspacing

\bibitem{ros2010rep105}
\BIBentryALTinterwordspacing
W.~Meeussen. (2010, October) Coordinate frames for mobile platforms. [Online].
  Available: \url{https://ros.org/reps/rep-0105.html}
\BIBentrySTDinterwordspacing

\bibitem{foote2013tf}
T.~Foote, ``tf: The transform library,'' in \emph{2013 IEEE Conference on
  Technologies for Practical Robot Applications (TePRA)}.\hskip 1em plus 0.5em
  minus 0.4em\relax IEEE, 2013, pp. 1--6.

\bibitem{merkel2014docker}
D.~Merkel, ``Docker: lightweight linux containers for consistent development
  and deployment,'' \emph{Linux journal}, vol. 2014, no. 239, p.~2, 2014.

\bibitem{optitrack2022mocap}
\BIBentryALTinterwordspacing
{NaturalPoint, Inc. DBA OptiTrack}. (2022) Motion capture systems. [Online].
  Available: \url{https://optitrack.com/}
\BIBentrySTDinterwordspacing

\bibitem{farnham2021umbrella}
T.~Farnham, S.~Jones, A.~Aijaz, Y.~Jin, I.~Mavromatis, U.~Raza, A.~Portelli,
  A.~Stanoev, and M.~Sooriyabandara, ``Umbrella collaborative robotics testbed
  and iot platform,'' in \emph{2021 IEEE 18th Annual Consumer Communications \&
  Networking Conference (CCNC)}.\hskip 1em plus 0.5em minus 0.4em\relax IEEE,
  2021, pp. 1--7.

\bibitem{mouret201720}
J.-B. Mouret and K.~Chatzilygeroudis, ``20 years of reality gap: a few thoughts
  about simulators in evolutionary robotics,'' in \emph{Workshop "Simulation in
  Evolutionary Robotics", Genetic and Evolutionary Computation Conference
  (GECCO 2017)}.\hskip 1em plus 0.5em minus 0.4em\relax Berlin, Germany: ACM,
  2017.

\bibitem{pickem2016safe}
D.~Pickem, L.~Wang, P.~Glotfelter, Y.~Diaz-Mercado, M.~Mote, A.~Ames, E.~Feron,
  and M.~Egerstedt, ``Safe, remote-access swarm robotics research on the
  robotarium,'' \emph{arXiv preprint arXiv:1604.00640}, 2016.

\bibitem{beltrame2018engineering}
G.~Beltrame, E.~Merlo, J.~Panerati, and C.~Pinciroli, ``Engineering safety in
  swarm robotics,'' in \emph{Proceedings of the 1st International Workshop on
  Robotics Software Engineering}, 2018, pp. 36--39.

\bibitem{eder2021complete}
K.~I. Eder, W.-l. Huang, and J.~Peleska, ``Complete agent-driven model-based
  system testing for autonomous systems,'' \emph{arXiv preprint
  arXiv:2110.12586}, 2021.

\bibitem{khronos2011webgl}
{Khronos Group}, \emph{WebGL 1.0 Specification}, Khronos Group, 2011.

\bibitem{jones2020distributed}
S.~Jones, E.~Milner, M.~Sooriyabandara, and S.~Hauert, ``Distributed
  situational awareness in robot swarms,'' \emph{Advanced Intelligent Systems},
  vol.~2, no.~11, p. 2000110, 2020.

\bibitem{pitonakova2018information}
L.~Pitonakova, R.~Crowder, and S.~Bullock, ``Information exchange design
  patterns for robot swarm foraging and their application in robot control
  algorithms,'' \emph{Frontiers in Robotics and AI}, vol.~5, p.~47, 2018.

\bibitem{jones2019onboard}
S.~Jones, A.~F. Winfield, S.~Hauert, and M.~Studley, ``Onboard evolution of
  understandable swarm behaviors,'' \emph{Advanced Intelligent Systems},
  vol.~1, 2019.

\bibitem{dorigo2013swarmanoid}
M.~Dorigo, D.~Floreano, L.~M. Gambardella, F.~Mondada, S.~Nolfi, T.~Baaboura,
  M.~Birattari, M.~Bonani, M.~Brambilla, A.~Brutschy \emph{et~al.},
  ``Swarmanoid: a novel concept for the study of heterogeneous robotic
  swarms,'' \emph{IEEE Robotics \& Automation Magazine}, vol.~20, no.~4, pp.
  60--71, 2013.

\bibitem{champandard2007behavior}
A.~Champandard, ``Behavior trees for next-gen game {AI},'' in \emph{Game
  developers conference, audio lecture}, 2007.

\bibitem{ogren2012increasing}
P.~Ogren, ``Increasing modularity of {UAV} control systems using computer game
  behavior trees,'' in \emph{AIAA Guidance, Navigation and Control
  Conference}.\hskip 1em plus 0.5em minus 0.4em\relax Minneapolis, MN, USA:
  AIAA, 2012.

\bibitem{klockner2013behavior}
A.~Kl{\"o}ckner, ``Behavior trees for uav mission management.'' in
  \emph{INFORMATIK 2013 Informatik angepasst an Mensch, Organisation und
  Umwelt}.\hskip 1em plus 0.5em minus 0.4em\relax Koblenz, Germany: Springer,
  2013, pp. 57--68.

\bibitem{jones2016evolving}
S.~Jones, M.~Studley, S.~Hauert, and A.~F. Winfield, ``Evolving behaviour trees
  for swarm robotics,'' in \emph{13th International Symposium on Distributed
  Autonomous Robotic Systems (DARS 2016)}, R.~Gro{\ss}, A.~Kolling, S.~Berman,
  E.~Frazzoli, A.~Martinoli, F.~Matsuno, and M.~Gauci, Eds.\hskip 1em plus
  0.5em minus 0.4em\relax London, UK: Springer, 2016.

\bibitem{colledanchise2017behavior}
M.~Colledanchise, ``Behavior trees in robotics,'' 2017.

\bibitem{marzinotto2014towards}
A.~Marzinotto, M.~Colledanchise, C.~Smith, and P.~Ogren, ``Towards a unified
  behavior trees framework for robot control,'' in \emph{IEEE International
  Conference on Robotics and Automation (ICRA 2014)}.\hskip 1em plus 0.5em
  minus 0.4em\relax Hong Kong, China: IEEE, 2014, pp. 5420--5427.

\end{thebibliography}

%
%
%

%







\end{document}